\documentclass{article} % For LaTeX2e
\usepackage[preprint]{colm2026_conference}

\usepackage{microtype}
\usepackage{hyperref}
\usepackage{url}
\usepackage{wrapfig}
\usepackage{subcaption}
\usepackage{booktabs}
\usepackage{tcolorbox}
\usepackage{amsmath,amssymb,amsthm}
\usepackage{enumitem}
\usepackage{xspace}
\usepackage{graphicx}
\usepackage{fontawesome5}
\usepackage{lineno}

\definecolor{darkblue}{rgb}{0, 0, 0.5}
\hypersetup{colorlinks=true, citecolor=darkblue, linkcolor=darkblue, urlcolor=darkblue}

\title{Nonsense Helps: Prompt Space Perturbation Broadens Reasoning Exploration}

\author{Langlin Huang, Chengsong Huang, Jinyuan Li, Donghong Cai, Yuyi Yang, Jiaxin Huang \\
Washington University in St. Louis\\
\texttt{\{h.langlin, jiaxinh\}@wustl.edu}
}

\newcommand{\method}{\textsc{\textbf{LoPE}}\xspace}

\begin{document}

\ifcolmsubmission
\linenumbers
\fi

\maketitle

\begin{abstract}
Reinforcement learning with verifiable rewards, particularly Group Relative Policy Optimization (GRPO), has significantly advanced the reasoning capabilities of Large Language Models (LLMs). However, in complex tasks, GRPO frequently suffers from the ``zero-advantage problem'': when all sampled rollouts for a query fail, the relative advantage collapses to zero. Consequently, the model loses effective training signals for these questions, wasting the training data and computational budget. While simply increasing the sampling budget for these questions is a common remedy, the static sampling policy inherently constrains reasoning exploration, limiting the success rate. In this paper, we propose Lorem Perturbation for Exploration (\method), a simple yet effective training framework to break this exploration bottleneck. We posit that task-irrelevant prompt-space perturbations can shift the model's output distribution enough to unlock orthogonal reasoning pathways for hard questions. Specifically, \method prepends sequences stochastically assembled from Lorem Ipsum vocabulary (a pseudo-Latin placeholder text) to the prompts before resampling. Experiments across 1.7B, 4B, and 7B models demonstrate that \method significantly outperforms resampling with the original prompts. Further analysis reveals that other Latin-based random sequences with low perplexity are also effective perturbations. Our results establish \method as a strong baseline for broadening exploration in LLM reinforcement learning. 
\end{abstract}
\vspace{-1.5em}
\begin{center}
\faGithub\ Code: \href{https://github.com/shrango/LoPE}{\texttt{https://github.com/shrango/LoPE}}
\end{center}
\vspace{-1.5em}
\section{Introduction}
In recent years, Reinforcement Learning with Verifiable Rewards (RLVR) has proven highly effective in enhancing the reasoning capabilities of large language models (LLMs). Notably, Group Relative Policy Optimization (GRPO)~\citep{yang2024qwen25mathtechnicalreportmathematical, Guo_2025Deepseekr1} has been widely recognized as a promising method. By leveraging the relative advantages among multiple responses generated for the same query, GRPO eliminates the need for a separate value model. However, this approach is severely compromised by the ``zero-advantage problem'': when all sampled responses to a question fail, their relative advantages collapse to zero. As a result, the vital training signal for that query is lost, wasting not only valuable training data but also a massive computational cost during LLM rollouts.

A simple solution to this problem is to generate more responses per question. To achieve this, many works have explored adaptive rollout budget allocation~\citep{liao2025enhancingefficiencyexplorationreinforcement, li2025knapsackrlunlockingexploration, xiong2025reinforceadaadaptivesamplingframework}. By providing more sampling attempts to hard questions, LLM has a better chance of hitting a correct answer and recovering the lost training signal. However, this approach has a clear limitation. Because these questions are simply too difficult for the model's current policy, merely increasing the sampling budget could still yield a low resample success rate.

Prior research has widely shown that modifications to the input context implicitly influence an LLM's output distribution~\citep{xie2022explanationincontextlearningimplicit, dai2023gptlearnincontextlanguage, goldwaser2025equivalencecontextparameterupdates}. Building on this principle, we hypothesize that a deliberate, prompt-level perturbation could alter the output distribution just enough to rescue the model from the zero-advantage trap. If the model is persistently failing on a hard question, perturbing the prompt during the rollout phase might unlock orthogonal reasoning pathways and discover successful trajectories that standard resampling cannot reach.

To test this hypothesis without introducing misleading facts or task-relevant hints, we require a task-irrelevant perturbation. Therefore, we draw inspiration from \textit{Lorem Ipsum}, a pseudo-Latin placeholder text designed to mimic natural language without conveying actual semantic meaning. Specifically, we construct a perturbation by randomly sampling words from the \textit{Lorem Ipsum} vocabulary. By prepending the random \textit{Lorem Ipsum} to the standard prompt, we introduce a pure prompt-space perturbation. We refer to these modified inputs as Lorem-perturbed prompts.
Based on this insight, we propose Lorem Perturbation for Exploration (\method), a simple yet highly effective rollout-and-resample framework designed to address the zero-advantage issue. We find that resampling with Lorem-perturbed prompts achieves a higher success rate on previously failed questions. This improvement is consistently observed throughout the entire RLVR training process and ensures effective training signals on a broader set of training questions than repeatedly sampling with the original unmodified prompt.

\begin{figure}
    \centering
    \includegraphics[width=\linewidth]{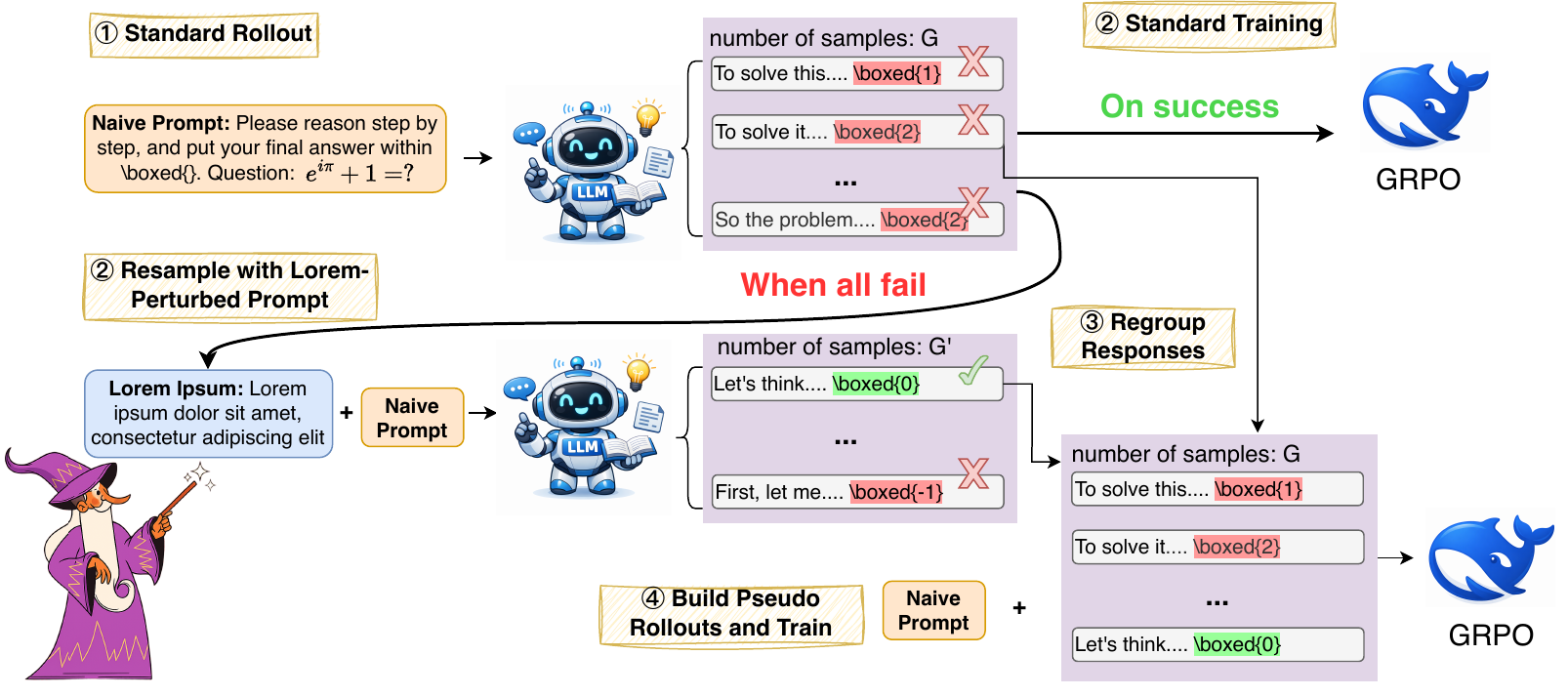}
    \caption{Overview of \method. During the standard rollout phase, if all $G$ responses fail, \method prepends a random \textit{Lorem Ipsum} sequence to the prompt and resamples $G'$ responses. Successful reasoning responses are regrouped with original failed responses to form a mixed batch of size $G$ for policy update.}
    \label{fig:method}
    \vspace{-1.5em}
\end{figure}

\method follows a similar training procedure to GRPO but differs in how it handles zero-advantage cases. For questions where all initial responses fail, we resample using Lorem-perturbed prompts instead of the naive prompt. 
% Experimental results show that Lorem-perturbed prompts achieve a substantially higher resample success rate compared to naive prompts.
Experimental results show that our method consistently improves model performance across multiple mathematical reasoning benchmarks, achieving an average gain of +2.79 points on Qwen3-1.7B-Base, +4.62 points on Qwen3-4B-Base, and +6.20 points on Qwen2.5-Math-7B. 

Furthermore, we conduct a comprehensive comparison of various prompt-space perturbation methods. While not all random perturbation strategies yield substantial improvements as \method does, the success of \method is not an isolated case. A few other perturbations, such as random sequences composed of high-frequency Latin words, achieve comparable results. We observe that the most effective perturbations share two decisive characteristics: (1) they use pseudo-Latin vocabularies to prevent interference with the English reasoning context, and (2) they maintain low perplexity to ensure high-quality rollouts. Overall, our results demonstrate that \method serves as a strong and generalizable baseline for broadening exploration in LLM reinforcement learning.

\section{Background: Group Relative Policy Optimization (GRPO)}
Group Relative Policy Optimization (GRPO)~\citep{shao2024deepseekmathpushinglimitsmathematical} is a widely used reinforcement learning algorithm for improving LLM reasoning capabilities. Compared with PPO-based approaches~\citep{schulman2017proximalpolicyoptimizationalgorithms}, GRPO is free of an explicit reward model and leverages the relative correctness among multiple responses sampled for the same question.

Formally, given a query $q$ and prompt $p$, it samples a group of \( G \) responses \( \{o_i\}_{i=1}^{G} \) from the old policy \( \pi_{\theta_{\text{old}}} \), where each response is a sequence \( o_i = (o_{i,1}, \dots, o_{i,|o_i|}) \). 

The training objective is to maximize the following equation:

\begin{equation}
\resizebox{\linewidth}{!}{$\displaystyle
J_{\text{GRPO}}(\theta)
= \mathbb{E}_{q, \{o_i\}_{i=1}^{G} \sim \pi_{\theta_{\text{old}}}(O|p,q)}
\frac{1}{G} \sum_{i=1}^{G} \frac{1}{|o_i|}
\sum_{t=1}^{|o_i|}
\Bigg\{
\min \Bigg[
\rho_{i,t} A_{i},
\operatorname{clip}\big(
\rho_{i,t}, 1-\epsilon, 1+\epsilon
\big)A_{i}
\Bigg]
- \beta D_{\mathrm{KL}}\!\left[\pi_\theta \,\|\, \pi_{\text{ref}}\right]
\Bigg\}
$},
\label{eq:grpo}
\end{equation}

where $\rho_{i,t}$ is the importance sampling ratio, defined as: 
% $    \rho_{i,t}=\frac{\pi_\theta(o_{i,t} \mid p,q, o_{i,<t})}
% {\pi_{\theta_{\text{old}}}(o_{i,t} \mid p,q, o_{i,<t})}$.
\begin{equation}
    \rho_{i,t}=\frac{\pi_\theta(o_{i,t} \mid p,q, o_{i,<t})}
{\pi_{\theta_{\text{old}}}(o_{i,t} \mid p,q, o_{i,<t})}.
\end{equation}
Here, \( \pi_\theta \) is the current policy, \( \pi_{\theta_{\text{old}}} \) is the old policy used for sampling. \( \pi_{\text{ref}} \) is the reference policy that serves as a regularizer to prevent \( \pi_\theta \) from deviating excessively from the initial distribution. This is achieved by a Kullback-Leibler (KL) divergence term, where $\beta$ controls the weight of KL. The clipping parameter \( \epsilon \) prevents excessively large policy updates that could destabilize training.

Let $r_i$ denote the scalar reward of the $i^\mathrm{th}$ response, where $i\in\{1,2,...,G\}$. The rollout-level advantage \( A_{i} \) is computed by normalizing the rewards within the same group: 
% $A_{i} = \dfrac{r_i-\mathrm{mean}(\textbf{r})}{\mathrm{std}(\textbf{r})}$. 
\begin{equation}
\label{eq:standard_adv}
    A_{i} = \dfrac{r_i-\mathrm{mean}(\textbf{r})}{\mathrm{std}(\textbf{r})}, \quad \mathbf{r} = [r_1, \ldots, r_G].
\end{equation}
Particularly, when all sampled responses to a question fail, resulting in a zero reward vector ($\textbf{r}=0$), the advantage $A_{i}$ collapses to $0$ for all $i$. Consequently, the training batch yields a zero gradient, wasting the computational budget allocated for the rollouts.

\section{The Limitation of Logit-Space Exploration}
\label{sec:prelim}

\begin{wrapfigure}{r}{0.7\textwidth}
    \centering
    
    \begin{subfigure}{0.48\linewidth}
        \centering
        \includegraphics[width=\linewidth]{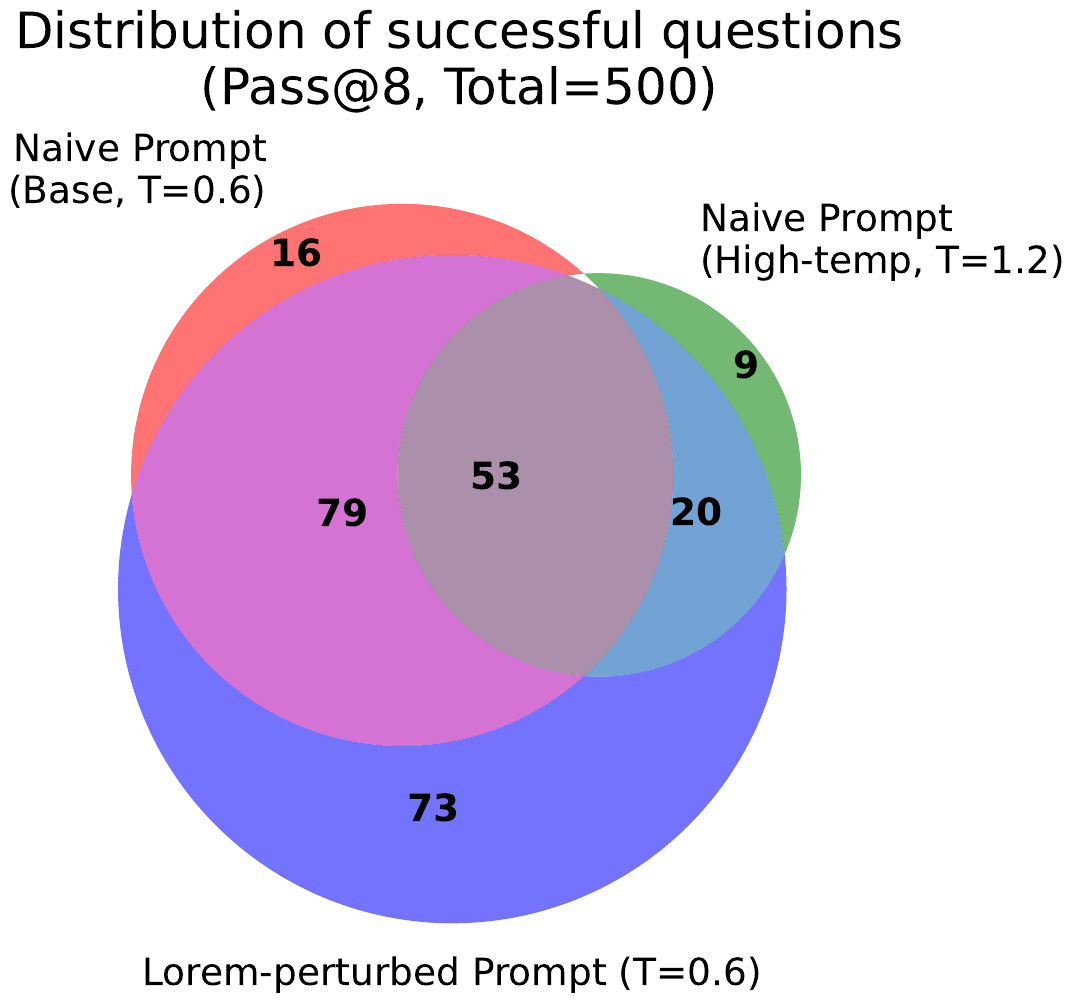}
        \caption{500-question subset}
        \label{subfig:venn_naive}
    \end{subfigure}
    % \hfill
    \begin{subfigure}{0.465\linewidth}
        \centering
        \includegraphics[width=\linewidth]{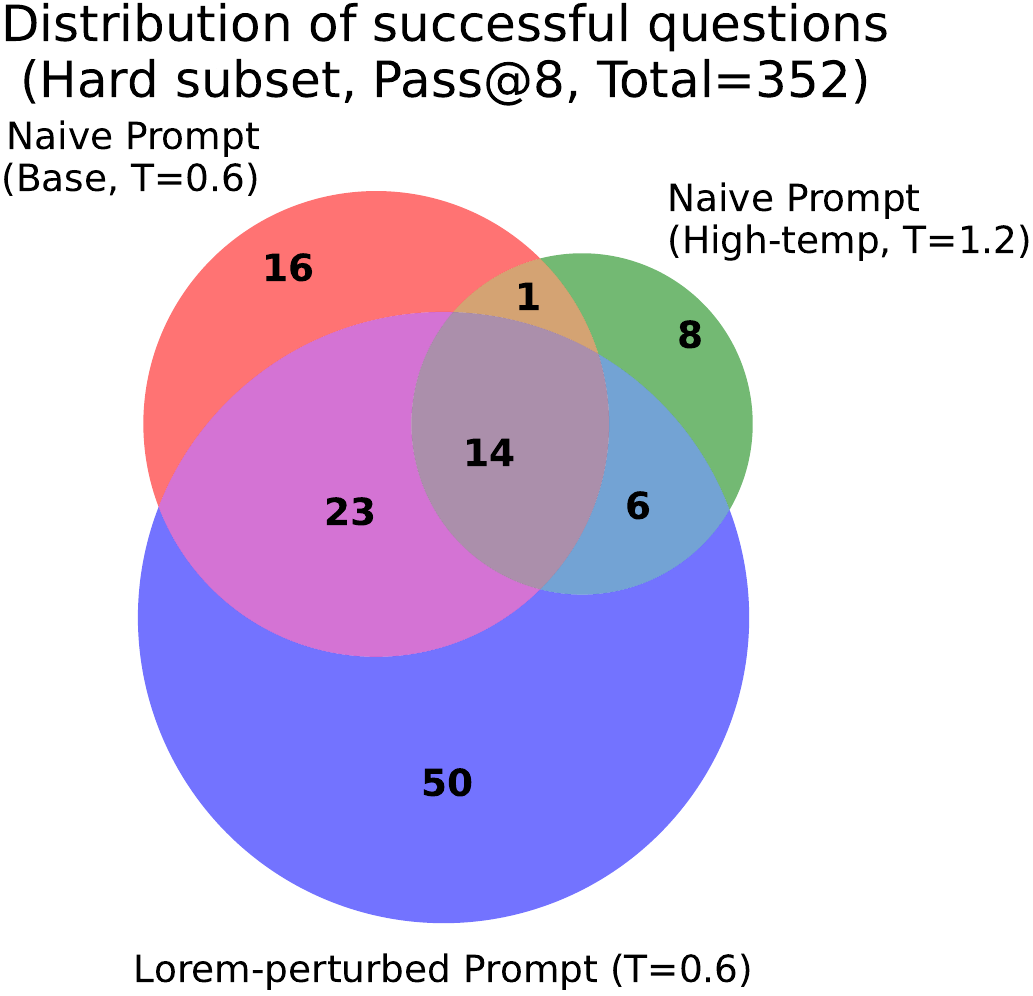}
        \caption{Hard 352-question subset}
        \label{subfig:venn_hard}
    \end{subfigure}

    \caption{Venn diagrams of successfully resolved questions (Pass@8) between naive prompting, Lorem perturbation, and high-temperature settings.}
    \label{fig:venn_combined}
    \vspace{-2em}
\end{wrapfigure}

When GRPO encounters the zero-advantage issue, a common and straightforward remedy is to resample additional responses for those questions. However, if an LLM fails to produce any correct answer within the first $G$ rollouts (e.g., $G=8$), it indicates that the question is intrinsically difficult under the current generation policy. In such cases, standard resampling is unlikely to significantly improve the resample success rate.

Traditionally, LLM generation encourages exploration by operating in the logit space, such as high-temperature sampling. We hypothesize that using high-temperature sampling alone is insufficient to shift the model out of its local reasoning basin.
Previous work extensively studied that In-Context Learning (ICL) is essentially changing the model's output distribution~\citep{xie2022explanationincontextlearningimplicit}. In this paper, we investigate whether \textbf{prompt-space perturbation}, which perturbs the input context, can more effectively force the model to explore orthogonal reasoning trajectories compared to \textbf{logit-space perturbation}. 

To this end, we conduct a preliminary experiment comparing three settings: (1) \textbf{Naive Prompt (Base)}: The original prompt with the system prompt and question only with a standard evaluation temperature of 0.6, serving as the base setting, (2) \textbf{Naive Prompt (High-temp)}: the original prompt with a higher temperature of 1.2 to encourage greater logit-space exploration, and (3) \textbf{Lorem-perturbed Prompt}: we prepend a randomly generated \textit{Lorem Ipsum} sequence to the naive prompt while keeping the temperature at 0.6.

\textbf{Lorem Ipsum} is a standard placeholder text widely used in publishing and graphic design. It consists of meaningless pseudo-Latin text that mimics the typical structural and statistical properties of natural language (such as word lengths and sentence boundaries) without carrying any meaningful semantic content. We use the \texttt{python-lorem} implementation~\citep{pypilorem}, where each word is uniformly sampled from a pool of 63 Latin words.

\begin{wrapfigure}{r}{0.7\textwidth}
    \centering
    \vspace{-1em}
    \includegraphics[width=\linewidth]{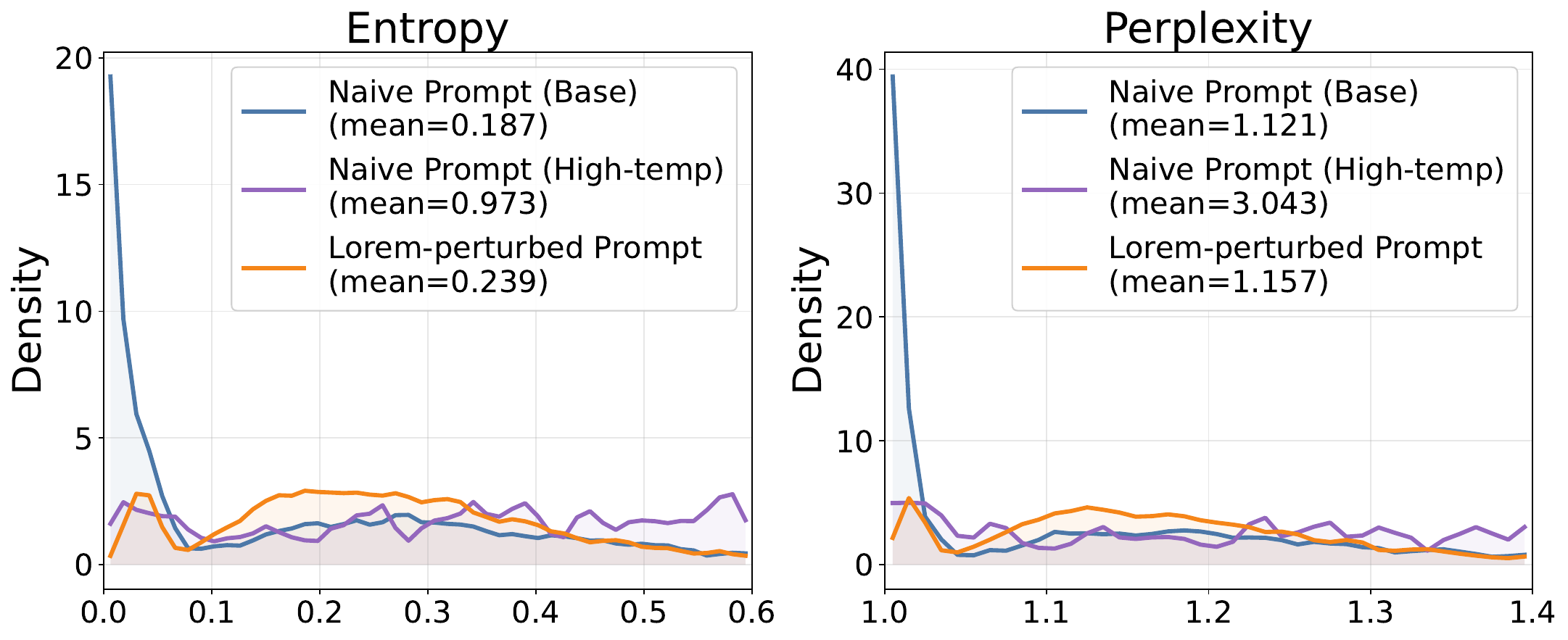}
    \caption{Probability distributions of response entropy and perplexity across different prompting and sampling formulations.}
    \label{fig:preli-entropy-perplexity}
    \vspace{-0.5em}
\end{wrapfigure}

We evaluate these prompt formulations on 500 randomly sampled questions from the Openr1-Math-46k-8192 dataset~\citep{yan2025learningreasonoffpolicyguidance}, using the Qwen3-1.7B-Base model~\citep{qwen3technicalreport}. To visually quantify the exploration overlap among different prompting strategies, we plot Venn diagrams (Figure~\ref{fig:venn_combined}) to show the set of distinct and overlapping questions successfully resolved within Pass@8 by each prompting formulation. Results for the 500-question evaluation set are shown in Figure~\ref{subfig:venn_naive}.
Furthermore, we construct a hard subset consisting of 352 questions that fail under the initial Pass@8 under the naive prompt setting. We then re-evaluate all three prompting formulations with a secondary 8-rollout sample budget on this subset, with results presented in Figure~\ref{subfig:venn_hard}.

Our observations are twofold. First, the Lorem-perturbed generations can actually resolve a large number of questions compared to both standard logit-space approaches (base and higher temperature).
Second, prompt-space perturbation unlocks orthogonal reasoning spaces that logit-space methods fail to explore. As shown in Figure~\ref{subfig:venn_hard}, when resampling the 352 hard questions,  the Lorem-perturbed responses independently resolve 50 unique questions that neither of the other two methods could answer.
This suggests that prompt-space perturbation can effectively broaden the exploration of LLM without degrading its overall reasoning ability, particularly on more challenging questions.

To further understand this phenomenon, we analyze the generated responses at the token level. Figure~\ref{fig:preli-entropy-perplexity} presents the probability density distributions of token-level entropy and perplexity across all responses. 
We observe that responses from the naive prompt (base) setting are heavily concentrated around low-entropy (near 0) and low-perplexity (near 1) regions, indicating highly confident but potentially over-constrained generation. In contrast, Lorem-perturbed responses eliminate the near-zero entropy spike and slightly right-shift the distribution, reflecting higher uncertainty and exploring behavior during generation. 
The high temperature setting, however, produces many responses with much higher entropy and perplexity, which can hurt the reasoning quality and accuracy. 

\section{Lorem Perturbation for Exploration (LoPE)}
\label{sec:lope}
Inspired by our findings that Lorem-perturbed prompts recover more failed questions and improve LLM exploration, we propose Lorem Perturbation for Exploration (\method), a simple yet effective resampling strategy to enhance exploration in reinforcement learning. We describe the details below.

\paragraph{Rollout with Perturbation.}
During the rollout stage, resampling is triggered only for questions where all $G$ responses $\{o_j\}_{j=1}^{G}$ generated from $\pi_{\theta_{\text{old}}}(o \mid p, q)$ under the naive prompt $p$ fail. For such cases, \method prepends a random \textit{Lorem Ipsum} sequence to the original prompt, serving as a text perturbation $\delta$. This results in a perturbed prompt $\delta \oplus p$, which is then used to generate an additional set of $G'$ responses: $\{o'_j\}_{j=1}^{G'} \sim \pi_{\theta_{old}}(o'|\delta \oplus p, q)$. An illustration of this process is presented in Figure~\ref{fig:method}.

\paragraph{Regroup LLM Responses.}
In the policy update stage, \method maintains the group size of $G$ for advantage calculation. Specifically, we construct a hybrid batch of rollouts by combining failed responses from the original rollouts with successful responses from the resampled set. Let $c$ denote the number of correct responses in the resampled set $\{o'_j\}_{j=1}^{G'}$. We randomly select $N_{s}=\min(c, G-1)$ correct responses from the resampled pool and use them to replace an equal number of failed responses in the original group. Importantly, we ensure that at least one incorrect response remains in the group, so that relative advantages are non-zero and meaningful for optimization.

\paragraph{Construct Pseudo Rollout with Resampled Responses.}
Directly grouping and comparing responses generated from different input contexts can cause a biased advantage estimation. To align the context of all responses for a given question, we construct pseudo rollouts by pairing each resampled response $o'$ with the naive prompt $p$ and question $q$ for training. 
Concretely, the full sequence used for training is $(p, q, o)$ for original rollouts and $(p, q, o')$ for resampled rollouts, despite $o'$ being generated under policy $\pi_{\theta_{\text{old}}}(o' \mid \delta \oplus p, q)$. 

This substitution, however, results in an off-policy optimization scenario. To correct for the discrepancy between the sampling and training policies, we apply the importance sampling ratio defined in Eq.~\eqref{eq:is_resample} for the resampled responses. 
% This adjustment ensures unbiased policy optimization while preserving the benefits of enhanced exploration.

\begin{equation}
\label{eq:is_resample}
\rho_{i,t} = \frac{\pi_\theta(o'_{i,t} \mid p, q, o'_{i,<t})}
{\pi_{\theta_{\text{old}}}(o'_{i,t} \mid \delta \oplus p, q, o'_{i,<t})}.
\end{equation}

\paragraph{Removal of KL Regularization.}
In addition, \method removes the KL regularization term in Eq.~\eqref{eq:grpo}. The introduction of random-word perturbations is intended to promote broader exploration, while a KL constraint restricts such distributional shifts and therefore counteracts this objective. Empirically, prior work~\citep{wang2026grouppatternselectionoptimization} has shown that removing KL regularization is beneficial when training with multiple prompt patterns.

\section{Training Signal Shaping}
\label{sec:shape}
Within the foundational \method framework, resampling via prompt space perturbation effectively enhances training data utilization. However, the off-policy training often diminishes the gradient magnitude of rare reasoning trajectories, as the policy probability $\pi_{\theta}$ in Eq.~\eqref{eq:is_resample} becomes small for these instances. Furthermore, while response regrouping concentrates computational resources on positive rollouts, calculating advantages solely based on the $G$ selected responses underestimates the difficulty of questions and reduces the advantage signals for rare success samples. 

To address these limitations, we introduce \textit{training signal shaping}, which incorporates a policy shaping strategy and an advantage shaping strategy. These components are specifically designed to mitigate the issues stemming from the importance sampling ratio $\rho_{i,t}$ and the biased advantage estimation, respectively.

\paragraph{Policy Shaping.} 
Training on pseudo-rollouts inherently constitutes an off-policy process due to the distributional discrepancy between the training policy $\pi_\theta(o'_{i,t} \mid p, q, o'_{i,<t})$ and the sampling policy $\pi_{\theta_{\text{old}}}(o'_{i,t} \mid \delta \oplus p, q, o'_{i,<t})$. Consequently, tokens with relatively low probabilities under $\pi_{\theta}$ suffer from suppressed training weights~\citep{wang2025aspoasymmetricimportancesampling}. To address this issue, we adapt the policy shaping mechanism proposed by \citet{yan2025learningreasonoffpolicyguidance}:
\begin{equation}
f(\rho_{i,t}) = \frac{\rho_{i,t}}{\rho_{i,t} + \gamma}~,
\end{equation}
where $\gamma$ is set to $0.1$. This function constraints the gradient magnitude for high-probability tokens while amplifying it for low-probability ones. This adjustment is particularly crucial for resampled responses, as critical reasoning steps are often assigned low probabilities under the naive policy and would otherwise be inappropriately underweighted during training. Notably, whereas \citet{yan2025learningreasonoffpolicyguidance} assumes $\pi_{\theta_{\text{old}}} \equiv 1$, \method utilizes the exact values of $\pi_{\theta_{\text{old}}}$. A detailed analysis of how policy shaping impacts the training when relaxing the assumption of a fixed $\pi_{\theta_{\text{old}}}$ is provided in Appendix~\ref{appendix:policyshape_math}.

\paragraph{Advantage Shaping.}
In standard GRPO, the advantage for each response is computed by normalizing rewards within the sampled group of $G$ responses, as defined in Eq.~\eqref{eq:standard_adv}. In our resampling setting, however, the $G$ responses selected for training comprise a mixture of original failed rollouts and resampled successful ones. Critically, the $G'$ discarded responses consist almost exclusively of failed attempts. Consequently, calculating the advantage solely on the $G$ selected responses underestimates the difficulty of the question. This underestimation suppresses the absolute value of positive advantages, subsequently reducing the training weight assigned to rare successful samples. 

To mitigate this bias, we propose an advantage shaping mechanism that computes the advantage over the complete set of $G + G'$ responses:
\begin{equation}
\label{eq:lope_adv}
\hat{A}_{i} = \frac{r_i - \mathrm{mean}(\mathbf{r}_{\text{all}})}{\mathrm{std}(\mathbf{r}_{\text{all}})}, \quad \mathbf{r}_{\text{all}} = [r_1, \ldots, r_G, r'_1, \ldots, r'_{G'}],
\end{equation}
while restricting the gradient updates to the $G$ selected responses. This formulation ensures that the normalization statistics faithfully reflect the true question difficulty, thereby restoring the authentic advantage values of successful samples and appropriately amplifying the reward signals for the rare successes. We quantitatively analyse the effect of advantage shaping in Appendix~\ref{appendix:advshape_math}, which amplifies the positive advantages by a factor of 2.1 to 5.0.

\paragraph{Full Training Objective.}
Combining the components above, the complete training objective of \method is formulated as:
\begin{equation}
\begin{split}
J_{\text{\method}}(\theta)
= \frac{1}{G} \Bigg\{
&\mathbb{E}_{q, \{o_i\}_{i=N_{s}+1}^{G} \sim \pi_{\theta_{\text{old}}}(O|p,q)}
\sum_{i=N_{s}+1}^{G} \frac{1}{|o_i|}
\sum_{t=1}^{|o_i|}
\min \left[
\rho_{i,t} \hat{A}_{i},\
\operatorname{clip}\big(
\rho_{i,t}, 1-\epsilon, 1+\epsilon
\big)\hat{A}_{i}
\right] \\
+
&\mathbb{E}_{q, \{o_i\}_{i=1}^{N_{s}} \sim \pi_{\theta_{\text{old}}}(O|\delta \oplus p,q)}
\sum_{i=1}^{N_{s}} \frac{1}{|o_i|}
\sum_{t=1}^{|o_i|}
\Big[
f(\rho_{i,t}) \hat{A}_{i}
\Big]
\Bigg\}
\end{split}
\label{eq:lope}
\end{equation}
where the first term corresponds to the standard GRPO updates on the original rollouts, and the second term incorporates policy shaping via $f(\rho_{i,t})$ for the resampled responses. The application of training signal shaping effectively resolves the limitations of the standard \method.

\section{Experiment}
\subsection{Experiment Setup}
\label{sec:exp_setup}
We evaluate \method on three base models: Qwen3-1.7B-Base, Qwen3-4B-Base~\citep{qwen3technicalreport}, and Qwen2.5-MATH-7B~\citep{yang2024qwen25mathtechnicalreportmathematical}. 
For Qwen2.5-MATH-7B, whose original context length is 4096, we follow \citet{yan2025learningreasonoffpolicyguidance} to extend the context window to 16384. During training, the maximum response length is set to 8192 tokens, and the maximum input length is 2048 tokens. Our implementation is based on EasyR1~\citep{zheng2025easyr1}. Experiments on the 1.7B and 4B models are conducted on 4 $\times$ 80GB A100 GPUs, and those on the 7B model are conducted on 8 $\times$ 80GB A100 GPUs.

We use the OpenR1-Math-46k-8192 dataset~\citep{yan2025learningreasonoffpolicyguidance} for training. For evaluation, we consider a diverse set of math reasoning benchmarks, including MATH-500~\citep{hendrycks2021measuringmathematicalproblemsolving}, GSM8K~\citep{cobbe2021trainingverifierssolvemath}, AMC (AMC 2022, 2023, and 2024), AIME 2024, and AIME 2025. We use EvalScope~\citep{evalscope_2024} for evaluation, with a sampling temperature of 0.6 and top-$p$ set to 0.95. For MATH-500, GSM8K, and AMC, we report Acc@1. For the more challenging benchmarks AIME 2024 and AIME 2025, we report Mean@32.

We compare \method against standard GRPO and the naive-prompt resampling baseline. The group size is set to $G=8$, and the resampling size is $G'=24$. All rollouts are performed with a default temperature of 1.0. For fair comparison, all resampling-based methods remove the KL regularization term.

For perturbation generation, Lorem Ipsum text is sampled using the \texttt{python-lorem} package\footnote{\url{https://pypi.org/project/python-lorem/}}. 
The sequence length is uniformly sampled between 100 and 300 tokens. 
Empirically, we append a short \textbf{boundary instruction} to the end of each perturbation sequence: \textit{“\textbackslash nPlease reason step by step, and put your final answer within \textbackslash boxed{}.”} This simple design effectively reduces cases in which the perturbation negatively interferes with the model and causes it to generate corrupted outputs, like random symbols and characters.

\subsection{Main Results}
\begin{table}[htbp]
\centering
\label{tab:main_results}
\resizebox{\textwidth}{!}{
\begin{tabular}{l cccccc}
\toprule
\textbf{Model \& Method} & \textbf{MATH-500} & \textbf{GSM8K} & \textbf{AMC} & \textbf{AIME24} & \textbf{AIME25} & \textbf{Avg.} \\ 
\midrule
\textbf{Qwen3-1.7B-Base}     & 63.40 & 76.92 & 26.87 & 5.33 & 2.00 & 34.90 \\
\quad + GRPO                     & 64.20 & 82.71 & 27.61 & 6.15 & 4.47 & 37.03 \\ 
\multicolumn{7}{l}{\qquad \textit{+ Resample}} \\
\qquad \hspace{0.5em} w/ Naive Prompt   & 67.00 & 82.18 & 28.36 & 8.70  & 4.58  & 38.16 \\
\qquad \hspace{0.5em} w/ \method (w/o Training Signal Shaping)  & 68.00 & \textbf{83.55} & \textbf{33.58} & 7.97  & \textbf{5.83}  & 39.79 \\ 
% \qquad \qquad + \hll{advantage shaping} & 61.80 & 79.53 & 27.61 & 4.79  & 3.96  & 35.54 \\ 
% \qquad \qquad \qquad + \hll{policy shaping} & \textbf{68.80} & 82.94 & 32.84 & \textbf{8.80}  & 5.73  & \textbf{39.82} \\ 
\qquad \hspace{0.5em} w/ \method (w/ Training Signal Shaping) & \textbf{68.80} & 82.94 & 32.84 & \textbf{8.80}  & 5.73  & \textbf{39.82} \\ 
\midrule
\textbf{Qwen3-4B-Base}         & 65.80 & 82.71 & 32.84 & 9.38 & 7.24 & 39.59 \\
\quad + GRPO                     & 77.80 & 91.74 & 47.76 & 16.41 & 13.12 & 49.37 \\ 
\multicolumn{7}{l}{\qquad \textit{+ Resample}} \\
\qquad \hspace{0.5em} w/ Naive Prompt   & 79.80 & 92.87 & 45.52 & 14.90 & 11.67 & 48.95 \\
\qquad \hspace{0.5em} w/ \method (w/o Training Signal Shaping) & \textbf{85.40} & \textbf{92.95} & 52.99 & 19.01 & 13.85 & 52.84 \\ 
% \qquad \qquad + \hll{advantage shaping} & 82.00 & 92.72 & 54.48 & 17.81  & \textbf{17.60}  & 52.92 \\ 
% \qquad \qquad \qquad + \hll{policy shaping} & 82.60 & \textbf{92.95} & \textbf{58.21} & \textbf{19.90} & 16.27  & \textbf{53.99} \\ 
\qquad \hspace{0.5em} w/ \method (w/ Training Signal Shaping) & 82.60 & \textbf{92.95} & \textbf{58.21} & \textbf{19.90} & \textbf{16.27}  & \textbf{53.99} \\ 
\midrule
\textbf{Qwen2.5-Math-7B}         & 52.80 & 65.50 & 35.40 & 12.90 & 7.90 & 34.90 \\
\quad + GRPO                     & 78.00 & 85.06 & 47.76 & 17.66 & 9.90 & 47.68 \\ 
\multicolumn{7}{l}{\qquad \textit{+ Resample}} \\
\qquad \hspace{0.5em} w/ Naive Prompt   & 78.20 & 83.02 & 50.00 & 17.19 & 9.17 & 47.52 \\
\qquad \hspace{0.5em} w/ \method (w/o Training Signal Shaping) & 77.40 & 86.35 & 47.01 & 15.31 & 10.52 & 47.32 \\ 
% \qquad \qquad + advantage shaping & 78.80 & 87.79 & 50.75 & 16.35 & 10.69  & 48.88 \\ 
% \qquad \qquad \qquad + policy shaping & \textbf{81.80} & \textbf{90.30} & \textbf{61.19} & \textbf{19.58} & \textbf{16.51}  & \textbf{53.88} \\ 
\qquad \hspace{0.5em} w/ \method (w/ Training Signal Shaping) & \textbf{81.80} & \textbf{90.30} & \textbf{61.19} & \textbf{19.58} & \textbf{16.51}  & \textbf{53.88} \\ 
\bottomrule
\end{tabular}
}
\caption{Performance comparison across different model scales. \method with training signal shaping consistently outperforms GRPO and resampling with the naive prompt baselines.}
\label{tab:main}
\vspace{-0.5em}
\end{table}

Table~\ref{tab:main} presents the main evaluation results of \method compared to standard GRPO and the naive resampling baseline.
\method improves the reasoning capabilities of the base models, yielding the highest average performance across the evaluated benchmarks. On the Qwen3-1.7B-Base model, \method achieves an average score of 39.79, outperforming standard GRPO by 2.76 points, and surpassing the Naive Prompt resampling baseline by 1.63 points. This demonstrates that expanding exploration via prompt-space perturbation is more effective than simply allocating more compute to do naive resampling for logit-space perturbation. Similarly, on the Qwen3-4B-Base model, \method outperforms standard GRPO by 3.47 points. Another finding is that Naive Prompt resampling actually degrades performance compared to standard GRPO, probably due to policy drift without KL regularization. However, \method discovers orthogonal, high-quality reasoning trajectories in resampling, therefore injecting highly variant responses that act as implicit regularizers. On the Qwen2.5-Math-7B model, naive resampling and \method perform similarly to standard GRPO, while \method with training signal shaping significantly outperforms GRPO by 6.20 points. This suggests that although \method increases the resample success rate, the gain is weakened by optimization inefficiency under off-policy update and biased advantage estimation. Training signal shaping alleviates this issue by amplifying learning signals on rare successful responses and informative low-frequency tokens.

\subsection{Successful Training-Time Exploration}
In Section~\ref{sec:prelim}, we show that \method improves resample accuracy. A natural question is whether this advantage persists throughout the RL training process. To investigate this, we track the resample accuracy of Qwen3-1.7B-Base during training, including both the question-level and response-level success rates, corresponding to pass@$G'$ and Mean@$G'$ respectively, where $G'=24$ is the resampling size.

\begin{figure}[h]
    \centering
    \includegraphics[width=\linewidth]{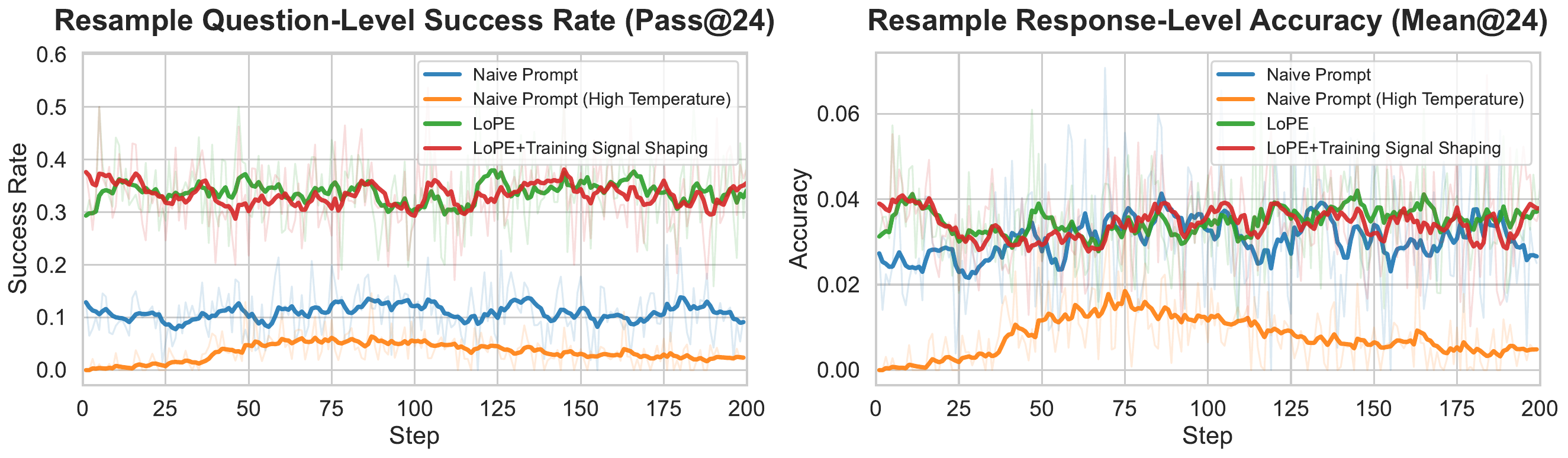}
    \caption{Resample success rate and accuracy of Qwen3-1.7B-Base during training.}
    \label{fig:training_curve}
    \vspace{-0.5em}
\end{figure}

We compare \method against two baselines: Naive prompt resampling and Naive prompt resampling with a high temperature of 1.2. The results are shown in Figure~\ref{fig:training_curve}. \method and \method with training signal shaping are similar in both subsets, indicating the training signal shaping does not alter rollout behavior. Interestingly, while \method and Naive prompt resampling achieve comparable performance in terms of response-level accuracy, \method consistently achieves a significantly higher question-level success rate. This indicates that the randomness introduced by \method enables the model to explore correct solutions across a broader set of questions, preventing the model from only optimizing on a narrow subset of training questions. Qwen3-4B-Base and Qwen2.5-Math-7B have similar observations, whose training-time accuracies are shown in Appendix~\ref{appendix:complete_training_dynamics}.

\begin{tcolorbox}[colback=blue!5!white, colframe=blue!60!black,boxsep=2pt,left=4pt, right=4pt, top=2pt, bottom=2pt]
\textbf{Takeaway}: \method maintains a higher resample success rate throughout the GRPO training process and ensures effective training signals on a broader set of questions.
\end{tcolorbox}
\vspace{-0.5em}

\section{Analysis: What Makes a Good Prompt Space Perturbation?}

In this section, we systematically investigate why Lorem Ipsum succeeds as a prompt space perturbation. We first propose a broad range of methods to generate random noise perturbations. Next, we conduct a comprehensive analysis of the properties of these perturbations. Finally, we perform RL training with these methods and conclude the underlying principles that define an effective prompt space perturbation.

\subsection{Exploring Alternative Prompt Space Perturbation}
We first introduce several methods for generating prompt space perturbations, primarily focusing on randomly generated noise:
\begin{itemize}[leftmargin=*, itemsep=0pt, topsep=0pt]
    % \item \textbf{Multi-style Prompt.} This setting uses system prompts of different styles as the perturbation, which are provided by~\citet{lu2026promptaugmentationscalesgrpo}. We adopt 12 of the 13 available prompt styles, excluding the one identical to our naive prompt.
    \item \textbf{Random Fake English.} This setting uses the \texttt{Faker} package\footnote{\url{https://pypi.org/project/Faker/}.} to generate fake English sentences, where each word is independently sampled from a commonly used English word pool\footnote{\url{https://github.com/joke2k/faker/blob/master/faker/providers/lorem/en_US/__init__.py}.}. In addition, it applies initial capitalization and inserts random punctuation to mimic natural sentence structures.
    \item \textbf{Random ASCII.} This setting uniformly samples printable ASCII characters to form random sequences.
    \item \textbf{Random Tokens.} This setting uniformly samples tokens from the model's vocabulary to form random sequences. Special tokens are excluded to prevent functional errors.
    \item \textbf{English Unigram Model.} In this setting, we follow the same random generation procedure as \method, but replace the candidate word pool with the top-50 most frequent words extracted from the English subset of the C4 multilingual corpus~\citep{JMLR:c4}\footnote{\url{https://huggingface.co/datasets/allenai/c4/viewer/la}}, which is widely used for pretraining. Non-English text is filtered out using \texttt{langid}\footnote{\url{https://pypi.org/project/langid/}} and \texttt{fastText}~\citep{joulin2016fasttext}. Words are sampled uniformly from the word pool.
    \item \textbf{Latin Unigram Model.} This setting is similar to \textbf{English Unigram Model}, but uses the top-50 most frequent Latin words as the candidate word pool.
    \item \textbf{Latin 3-Gram Model.} In this setting, we explore locally coherent random sequences generated by a 3-gram language model. Concretely, we train a 3-gram language model using \texttt{markovify}\footnote{\url{https://pypi.org/project/markovify/}} on the Latin subset of the C4 corpus, and use this model for random sequence generation during training.
    \item \textbf{Filtered Latin Natural Language.} In this setting, we use natural Latin text as the prompt space perturbation. Specifically, we adopt the Latin subset of the C4 corpus. We filter out non-Latin sentences using \texttt{langid}\footnote{\url{https://pypi.org/project/langid/}} and \texttt{fastText}~\citep{joulin2016fasttext}, and additionally remove sentences containing the canonical Lorem Ipsum incipit (\textit{``lorem ipsum dolor sit amet, consectetur adipiscing elit''}). After deduplication and retaining only sequences within the 100--300 token range, we obtain a corpus of approximately 65K sequences. 
\end{itemize}

The length of all prompt space perturbations is uniformly sampled between 100 and 300 tokens to match the main experiment.

\begin{figure}[h]
    \centering
    \includegraphics[width=\linewidth]{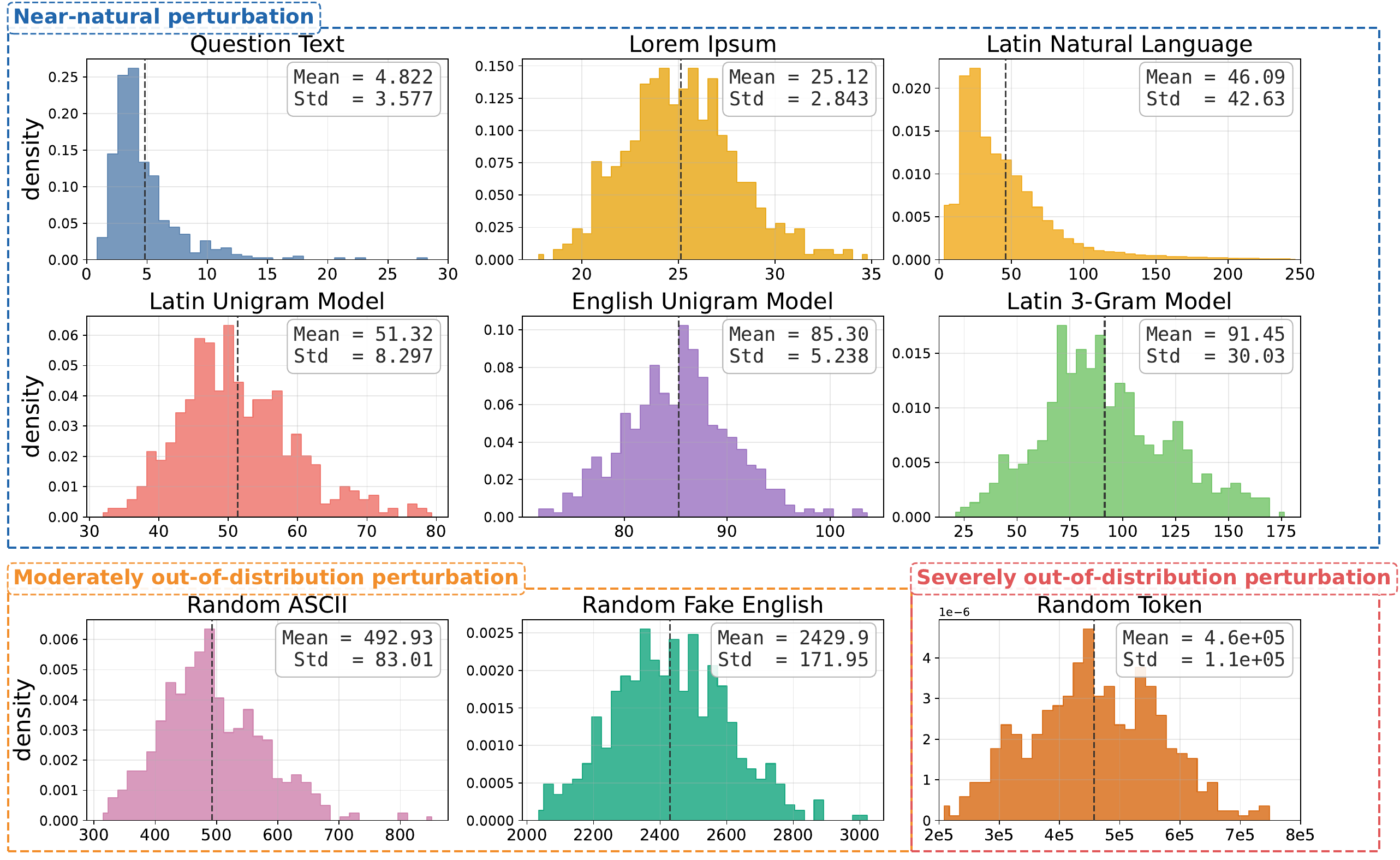}
    \caption{Perplexity distributions of randomly generated sequences from each perturbation type, with the mean and standard deviation reported in each subplot. ``Question Text'' from the 500-question evaluation set serves as an English natural language reference. Perturbations are grouped by their perplexity computed on Qwen3-1.7B-Base. Perturbations in the first two rows show a near-natural language property, yielding better final performance, while those in the last row are out of the model distribution, resulting in worse performance as shown in Table~\ref{table:random_method}.}
    \label{fig:random_seq_ppl}
\end{figure}

\subsection{How LLMs Perceive Prompt-Space Noise}
To understand how prompt space perturbations influence the model's generation process, we conduct a progressive three-step analysis. Specifically, we systematically investigate \textbf{(i)} the intrinsic properties of the perturbation text itself and \textbf{(ii)} its impact on the model's understanding of the question.
% , and (iii) its ultimate effect on the generated reasoning responses. 

\paragraph{How Does the LLM Understand the Random Perturbation Sequences?}
We begin by analyzing the intrinsic properties of different perturbation sequences, focusing on how well they align with the language model's distribution. For each method, we generate 500 perturbation sequences of 200 tokens each and compute their perplexity under Qwen3-1.7B-Base. As a reference for in-distribution natural language, we also report the perplexity of the question text from the 500-question evaluation subset.

The resulting distributions are presented in Figure~\ref{fig:random_seq_ppl}. We characterize each method's sequences by their \textit{mean} perplexity (indicating how far they deviate from natural language) and their \textit{standard deviation} (indicating the consistency of the perturbation strength across samples).

The mean perplexities span several orders of magnitude, naturally partitioning the methods into three regimes. \textbf{(i) Near-natural perturbations} (mean below 100): \textit{Lorem Ipsum} (25.12), \textit{Filtered Latin Natural Language} (46.09), \textit{Latin Unigram Model} (51.32), \textit{English Unigram Model} (85.30), and \textit{Latin 3-Gram Model} (91.45) all sit within roughly an order of magnitude of the \textit{Question Text} reference (4.82), suggesting they behave as structurally plausible noise from the model's perspective. \textbf{(ii) Moderately out-of-distribution perturbations} (mean in the hundreds to low thousands): \textit{Random ASCII} (492.93) drifts further from the language manifold, while \textit{Random Fake English} reaches a mean of 2429.9. \textbf{(iii) Severely out-of-distribution perturbations}: \textit{Random Token} explodes to a mean perplexity of $4.6 \times 10^5$, reflecting a complete collapse of linguistic structure.

Beyond the mean perplexity, the dispersion of these distributions also varies substantially. \textit{Lorem Ipsum} exhibits both the lowest mean value among the synthetic perturbations and a tightly concentrated distribution (std 2.84), indicating that every sampled sequence imposes a consistent, controlled distributional shift. In contrast, other low-mean perturbations exhibit a much wider spread. Most notably, \textit{Filtered Latin Natural Language} presents a long-tailed distribution (std 42.63) that extends beyond a perplexity of 200, despite having a comparable mean. Such uneven perturbation strength would inevitably lead to high within-batch variance during RL training. 
We therefore consider a filtered variant of \textit{Filtered Latin Natural Language} with sequences having perplexity between 20 and 30.

\begin{tcolorbox}[colback=blue!5!white, colframe=blue!60!black,boxsep=2pt,left=4pt, right=4pt, top=2pt, bottom=2pt]
\textbf{Takeaway}: Random sequences generated by sampling from a limited word pool have lower perplexity, showing a near-natural property.
\end{tcolorbox}

\paragraph{Impact of Random Perturbations on Question Comprehension.}
The previous analysis demonstrates that random sequences introduce varying degrees of perturbation into the prompt space, implicitly altering the model's policy. We next investigate the extent to which these perturbations disrupt the model's comprehension of the input, particularly the question text itself.

\begin{figure}[ht]
    \centering
    \begin{subfigure}{0.47\linewidth}
        \centering
        \includegraphics[width=\linewidth]{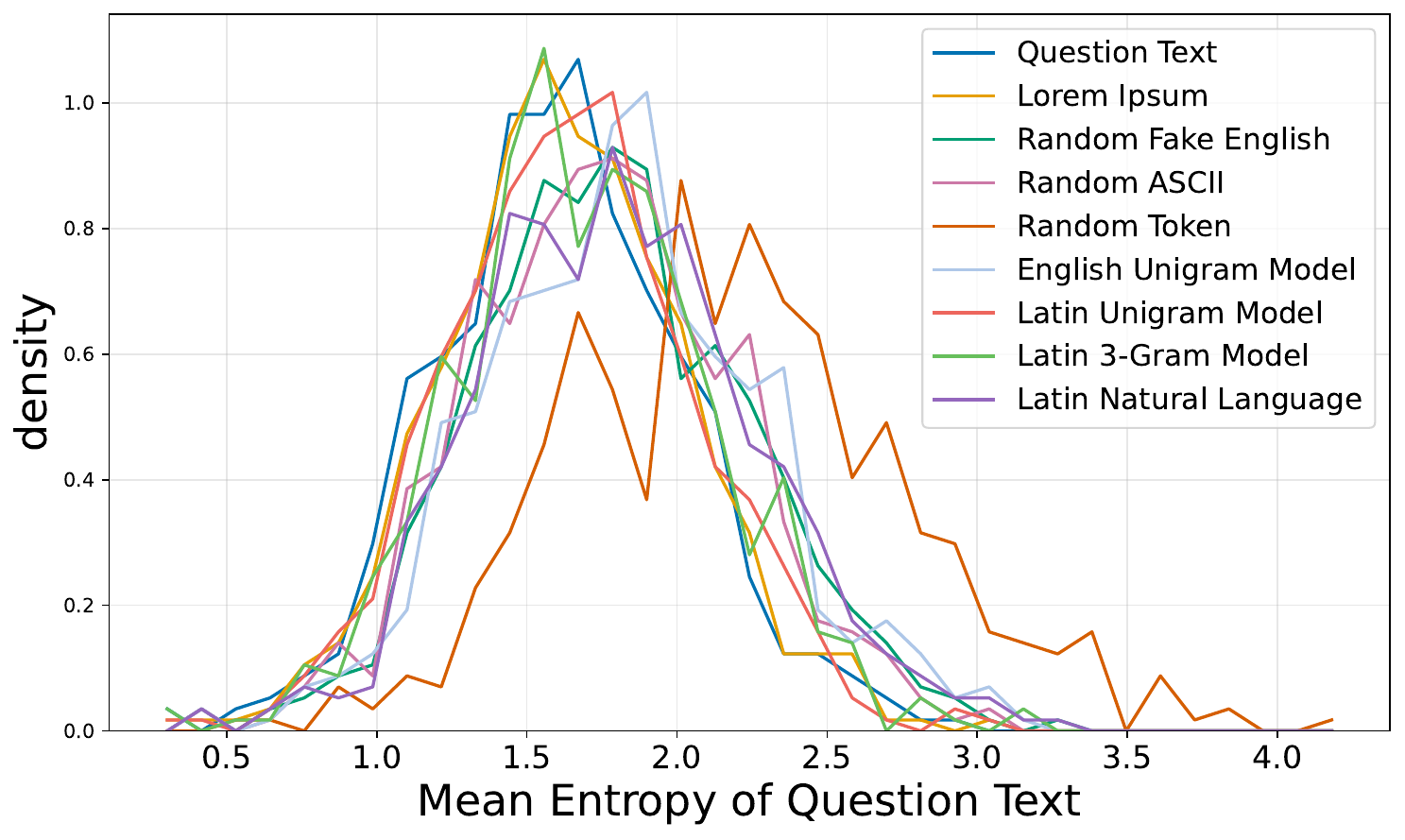}
        \caption{Entropy distribution of 500 question prompts. Most perturbations preserve a distribution close to the Question Text. The distribution of Random Token is clearly right-shifted, indicating a corrupted understanding of the question.}
        \label{subfig:input_prompt_entropy}
    \end{subfigure}
    \hfill
    \begin{subfigure}{0.47\linewidth}
        \centering
        \includegraphics[width=\linewidth]{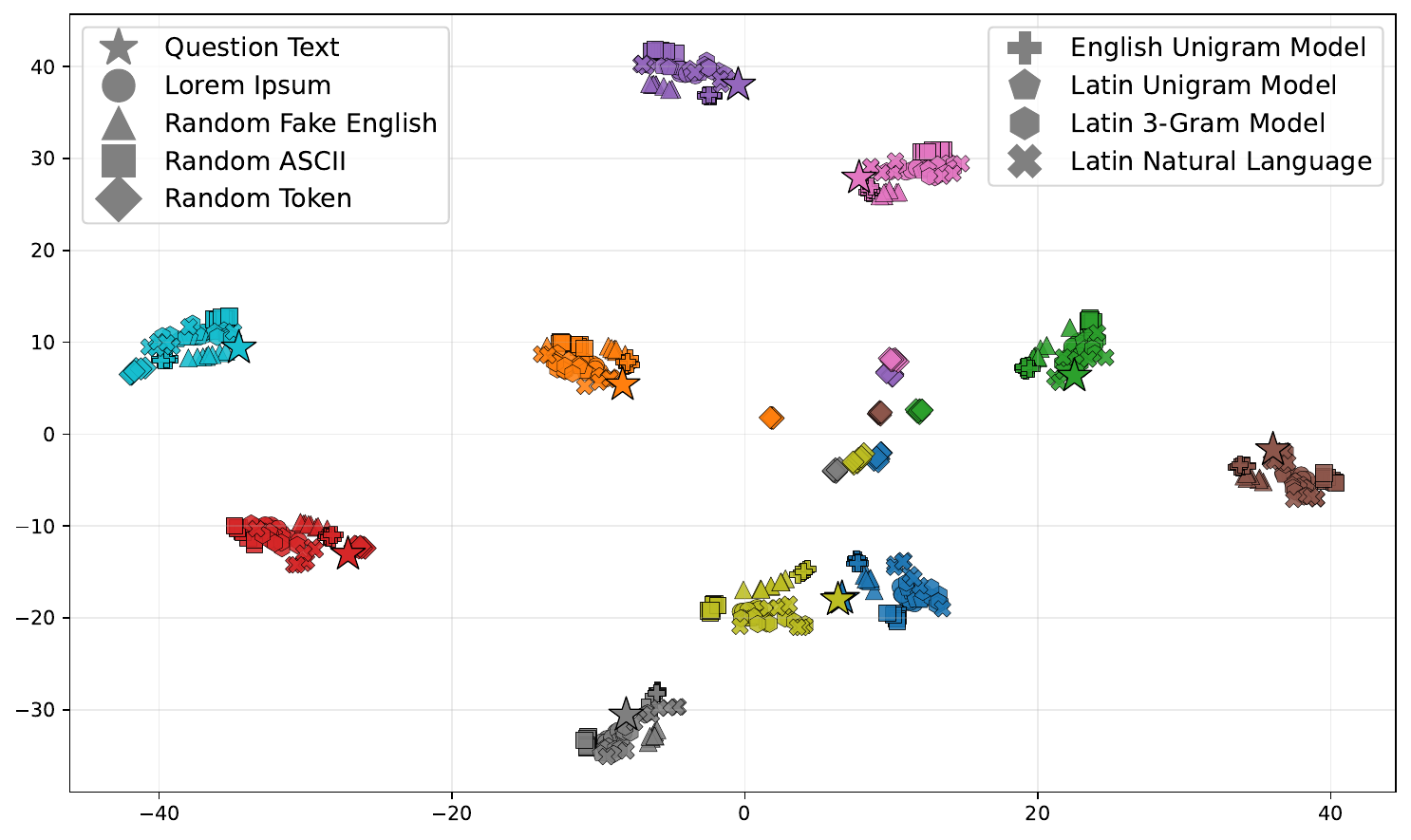}
        \caption{Question representation visualization via t-SNE under different perturbations. Each color corresponds to one question (8 samples per perturbation). Random Token perturbations drift far from the original meaning.}
        \label{subfig:input_prompt_sentence_representation}
    \end{subfigure}
    \vspace{-0.5em}
    \caption{The influence of various prompt space perturbations on question comprehension.}
    \label{fig:input_prompt_analysis}
\end{figure}

To achieve this, we analyze both the token-level entropy and the sentence-level representations of the questions. Figure~\ref{subfig:input_prompt_entropy} presents the entropy distribution across 500 question prompts, where each prompt is characterized by the average entropy of its constituent tokens. The entropy distributions of most perturbations largely overlap with the \textit{Question Text}, indicating that they alter LLM's question comprehension only slightly. In contrast, \textit{Random Tokens} exhibit a significant right-shifted deviation.

Figure~\ref{subfig:input_prompt_sentence_representation} illustrates the impact of various perturbations on the semantic representations of the question text. We select 10 questions that are successfully solved by all methods within 8 rollouts and compute their sentence representations, defined as the mean pooling of the final-layer hidden states across all tokens in the question. We then apply t-SNE for visualization. Points of the same color correspond to the identical underlying question, while different marker shapes indicate distinct prompt perturbation types.

We observe that the representations under near-natural or moderately out-of-distribution perturbations cluster tightly around those of the \textit{Question Text}, implying that the model maintains a consistent semantic understanding. Conversely, \textit{Random Token} perturbations consistently drift far from these clusters, suggesting that severe noise significantly distorts LLM's interpretation of the input question.

Collectively, these findings demonstrate that excessively high-perplexity perturbations can corrupt the model's understanding of the input question, thereby hindering its ability to discover correct solutions during the RL training process.

\begin{tcolorbox}[colback=blue!5!white, colframe=blue!60!black,boxsep=2pt,left=4pt, right=4pt, top=2pt, bottom=2pt]
\textbf{Takeaway}: Near-natural perturbation (low perplexity) successfully shifts the model's generation distribution without corrupting its semantic understanding of the core question. Conversely, excessively noisy perturbation destroys the input representation.
\end{tcolorbox}

\subsection{The Recipe for an Effective Perturbation}

We conduct experiments with these prompt space perturbation methods. For \textit{Filtered Latin Natural Language}, we observe in Figure~\ref{fig:random_seq_ppl} that the corpus contains many sequences with relatively low perplexity, which aligns with the characteristics of Lorem Ipsum. Therefore, we further filter the corpus to retain sequences with a perplexity between 20 and 30, resulting in approximately 38K sequences.

As a baseline for comparison, we select resampling with the naive prompt. Additionally, we experiment with resampling with a higher temperature, which encourages broader exploration in rollouts. 

% In addition to the random noise perturbations, we compare a Multi-style Prompt setting, which uses system prompts of different styles as the perturbation. This is inspired by~\citet{lu2026promptaugmentationscalesgrpo}, which provides 13 available prompt styles. We exclude the one identical to our naive prompt and adopt the rest 12. 

% Across all the perturbation methods, we apply the same training scheme as \method with training signal shaping. 

\begin{table}[htbp]
\centering
\resizebox{\columnwidth}{!}{%
\begin{tabular}{l cccccc}
\toprule
\textbf{Model \& Method} & \textbf{MATH-500} & \textbf{GSM8K} & \textbf{AMC} & \textbf{AIME24} & \textbf{AIME25} & \textbf{Avg.} \\ 
\midrule
\textbf{Qwen3-1.7B-Base}          & 63.40 & 76.92 & 26.87 & 5.33 & 2.00 & 34.90 \\
\quad + GRPO                       & 64.20 & 82.71 & 27.61 & 6.15 & 4.47 & 37.03 \\ 
\multicolumn{7}{l}{\qquad \textit{+ Resample w/o perturbation}} \\
\qquad \hspace{0.5em} w/ Naive Prompt             & 67.00 & 82.18 & 28.36 & 8.70  & 4.58  & 38.16 \\
\qquad \hspace{0.5em} w/ Naive Prompt (Temp=1.2)  & 64.40 & 82.87 & 31.34 & 8.65  & 4.48  & 38.35 \\
% \qquad \hspace{0.5em} w/ Multi-style Prompt  & 66.40 & 82.26 & 32.09 & 8.23  & 5.10  & 38.82 \\
\multicolumn{7}{l}{\qquad \textit{+ Resample w/ perturbation + Training Signal Shaping}} \\
\qquad \hspace{0.5em} w/ \method  & \textbf{68.80} & \textbf{82.94} & \textbf{32.84} & \textbf{8.80}  & \textbf{5.73}  & \underline{39.82} \\ 
% \qquad \hspace{0.5em} w/ Multi-style Prompt  & - & - & - & - & - & - \\
\qquad \hspace{0.5em} w/ Random Fake English             & 65.80 & 81.96 & 32.09 & 7.50 & 5.42  & 38.55 \\
\qquad \hspace{0.5em} w/ Random ASCII             & 66.20 & 82.94 & 28.36 & 8.12  & 5.32  & 38.19 \\
\qquad \hspace{0.5em} w/ Random Token             & 64.20 & 81.50 & 29.85 & 8.08  & 4.63  & 37.65 \\
\qquad \hspace{0.5em} w/ Filtered Latin Natural Language  & \textbf{68.80} & 82.71 & \textbf{32.84} & \textbf{9.32}  & 5.57  & \textbf{39.85} \\ 
\qquad \hspace{0.5em} w/ Latin Unigram Model  & \textbf{69.40} & \textbf{83.32} & 32.09 & 7.19  & \textbf{6.35}  & \underline{39.67} \\ 
\qquad \hspace{0.5em} w/ Latin 3-Gram Model  & 68.80 & 81.88 & 29.85 & 7.92 & 5.93  & 38.88 \\ 
\qquad \hspace{0.5em} w/ English Unigram Model & 67.00 & 83.32 & 28.36 & 8.49 & 5.42  & 38.52 \\ 
\bottomrule
\end{tabular}%
}\caption{Comparison of different prompt space perturbation methods. The three methods with the smallest perplexity value (\method, \textit{Filtered Latin Natural Language}, and \textit{Latin Unigram Model}) achieve the best performance. \textit{Random Token} with the highest perplexity even harms the training. This suggests that mild perturbations are sufficient to drive improvement while avoiding detrimental effects of excessive noise.}
\label{table:random_method}
\end{table}

The experiment results are presented in Table~\ref{table:random_method}, showing that performance varies across different methods. While not all candidates match the superior results of \method, \method is not the only effective approach. Specifically, \textit{Filtered Latin Natural Language} and \textit{Latin Unigram Model} both achieve average scores exceeding 39.6, suggesting that the success of prompt space perturbation is not isolated to the \textit{Lorem Ipsum} format. As illustrated in Figure~\ref{fig:random_seq_ppl}, these high-performing methods share two defining characteristics: they consist of Latin words and exhibit the lowest perplexity among all evaluated perturbations.

We conjecture the reason for their success as follows. Such perturbations provide sufficient signals to drive the rollout policy away from the naive policy, encouraging model exploration. Meanwhile, their relatively low perplexity avoids introducing excessive noise, thereby preserving the quality of the rollout responses.

Furthermore, we observe that \textit{English Unigram Model} underperforms \textit{Latin 3-Gram Model}, despite the former possessing a slightly lower average perplexity. This indicates that English-based perturbations are more prone to interfering with the model's original English reasoning context, which subsequently hinders performance.

\begin{tcolorbox}[colback=blue!5!white, colframe=blue!60!black,boxsep=2pt,left=4pt, right=4pt, top=2pt, bottom=2pt]
\textbf{Takeaway}: The most effective prompt space perturbation share these characteristics: (i) pseudo-Latin vocabularies to prevent interference with the English reasoning context, and (ii) maintaining low perplexity to ensure high-quality rollouts.
\end{tcolorbox}

\section{Related Work}
\vspace{-0.5em}
\paragraph{Zero-Advantage Recovery in RLVR.}
GRPO-style reinforcement learning~\citep{shao2024deepseekmathpushinglimitsmathematical,Guo_2025Deepseekr1,yu2025dapoopensourcellmreinforcement} improves LLM reasoning but suffers from zero-reward signals on hard prompts.
Recent efforts focus on improving rollout efficiency and recovering useful training signals, such as adaptive budget allocation, targeted exploration, scaffolded hints, and off-policy guidance~\citep{zhang2026trainlesslearnmore, le2026promptleftbehindexploiting, bamba2025xrpopushinglimitsgrpo, zhang2026scafgrposcaffoldedgrouprelative, yan2025learningreasonoffpolicyguidance, zhao2026selfdistilledreasoneronpolicyselfdistillation}.
Another line of work explores modifying prompt patterns or prefixes during training, such as prompt selection, augmentation, and prefix-level guidance~\citep{wang2026grouppatternselectionoptimization, lu2026promptaugmentationscalesgrpo, mundada2026wsgrpoweaklysupervisedgrouprelativepolicy}.
\vspace{-0.5em}
\paragraph{Context-level Perturbation.}
While in-context learning with logit-level arithmetics explicitly alters output distribution~\citep{huang-etal-2026-divide}, a group of work shows that prompt context can also implicitly influence the model’s generation behavior, where small changes can induce substantial shifts~\citep{xie2022explanationincontextlearningimplicit, vonoswald2023transformerslearnincontextgradient}.
Empirical studies on prompting further demonstrate that reasoning performance is highly sensitive to prompt format, demonstrations, and instruction structure~\citep{wei2023chainofthoughtpromptingelicitsreasoning, wang2023selfconsistencyimproveschainthought, huang2025efficient, zhou2023leasttomostpromptingenablescomplex, yang2024gpt, huang2026cats}.
More recent work highlights that adding synthetic or meaningless tokens can shift model activations and alter reasoning behavior~\citep{shi2025meaninglesstokensmeaningfulgains, gan2026neuralthicketsdiversetask}.

\section{Conclusion}
This paper proposes Lorem Perturbation for Exploration (\method), a simple yet effective prompt space perturbation method that prepends a randomly generated Lorem Ipsum sequence to the naive prompt. This introduces a controllable perturbation that enables LLM to explore alternative reasoning trajectories and succeed on previously failed questions. Moreover, resampling with LoPE during GRPO training significantly improves question-level success rates, enhances data utilization, and yields consistent performance gains.

Furthermore, a systematic comparison of various prompt space perturbation methods reveals that the success of \method is not an isolated case. Specifically, our analysis indicates that effective perturbations share two crucial characteristics: they are Latin-based and have relatively low perplexity. These findings highlight the importance of controlled perturbation in improving exploration for LLM reasoning.

\section*{Ethics Statement}
This work explores prompt-space perturbations using randomly generated sequences. As these sequences are generated automatically, we cannot guarantee that they are entirely free of potentially toxic, biased, or inappropriate words or expressions.

In addition, we observe that when the perturbation is excessively strong, the model may produce incoherent or nonsensical outputs that, in some cases, include undesirable or harmful content. This highlights a limitation of our approach: increased exploration via perturbation may come at the cost of reduced controllability over model outputs.

To mitigate these risks, our method emphasizes controlled perturbation, and our experiments suggest that moderate, language-like perturbations are less likely to negatively affect model behavior. Nevertheless, ensuring the safety and robustness of such perturbations remains an open challenge.

In future work, we plan to systematically study how different types and strengths of perturbations influence model behavior, with the goal of minimizing the risk of generating toxic or harmful content while preserving the benefits of improved exploration.

\bibliography{colm2026_conference}
\bibliographystyle{colm2026_conference}
\newpage

\appendix

\section{Example of Lorem Ipsum Prompt}
\label{appendix:example}
\begin{tcolorbox}[title=Lorem Ipsum Prompt]

\textbf{System}

in laboris velit ex est ex est exercitation sint ex enim ut incididunt tempor cupidatat nisi occaecat deserunt laboris quis voluptate qui consequat proident minim pariatur velit est aliquip do labore ut dolor et ullamco proident in non culpa est amet ipsum officia velit exercitation aute consectetur aliquip ullamco laboris laborum deserunt quis enim consectetur ea veniam officia commodo cupidatat culpa fugiat consequat eu enim mollit commodo enim eiusmod deserunt laborum ullamco minim irure consectetur quis ut nisi cillum adipiscing proident culpa id cupidatat ad amet sunt qui adipiscing mollit proident dolor pariatur minim eu quis ad excepteur nulla reprehenderit ea enim

Please reason step by step, and put your final answer within \textbackslash boxed\{\}.

\textbf{User}

Given \(abc \neq 0\), and \(a+b+c=0\). Then the value of the algebraic expression
\[
\frac{a^{2}}{bc}+\frac{b^{2}}{ac}+\frac{c^{2}}{ab}
\]
is:

\begin{itemize}
\item[(A)] 3
\item[(B)] 2
\item[(C)] 1
\item[(D)] 0
\end{itemize}

You FIRST think about the reasoning process as an internal monologue and then provide the final answer. The reasoning process MUST BE enclosed within \texttt{<think>} \texttt{</think>} tags. The final answer MUST BE put in \textbackslash boxed\{\}.

\textbf{Assistant}

\end{tcolorbox}

% \begin{tcolorbox}[title=Lorem Ipsum After Question Prompt]

% \textbf{System}
% Please reason step by step, and put your final answer within \textbackslash boxed\{\}.

% \textbf{User}

% Given \(abc \neq 0\), and \(a+b+c=0\). Then the value of the algebraic expression
% \[
% \frac{a^{2}}{bc}+\frac{b^{2}}{ac}+\frac{c^{2}}{ab}
% \]
% is:

% \begin{itemize}
% \item[(A)] 3
% \item[(B)] 2
% \item[(C)] 1
% \item[(D)] 0
% \end{itemize}

% in laboris velit ex est ex est exercitation sint ex enim ut incididunt tempor cupidatat nisi occaecat deserunt laboris quis voluptate qui consequat proident minim pariatur velit est aliquip do labore ut dolor et ullamco proident in non culpa est amet ipsum officia velit exercitation aute consectetur aliquip ullamco laboris laborum deserunt quis enim consectetur ea veniam officia commodo cupidatat culpa fugiat consequat eu enim mollit commodo enim eiusmod deserunt laborum ullamco minim irure consectetur quis ut nisi cillum adipiscing proident culpa id cupidatat ad amet sunt qui adipiscing mollit proident dolor pariatur minim eu quis ad excepteur nulla reprehenderit ea enim

% You FIRST think about the reasoning process as an internal monologue and then provide the final answer. The reasoning process MUST BE enclosed within \texttt{<think>} \texttt{</think>} tags. The final answer MUST BE put in \textbackslash boxed\{\}.

% \textbf{Assistant}

% \end{tcolorbox}

\section{Training-Time Resample Accuracy for Qwen3-4B-Base and Qwen2.5-Math-7B}
\label{appendix:complete_training_dynamics}
% \begin{figure}[h]
%     \centering
%     \includegraphics[width=\linewidth]{Figures/training_time_resample_acc_curve.pdf}
%     \caption{Resample success rate and accuracy during Qwen3-1.7B-Base training.}
%     \vspace{-0.5em}
% \end{figure}

\begin{figure}[h]
    \centering
    \includegraphics[width=\linewidth]{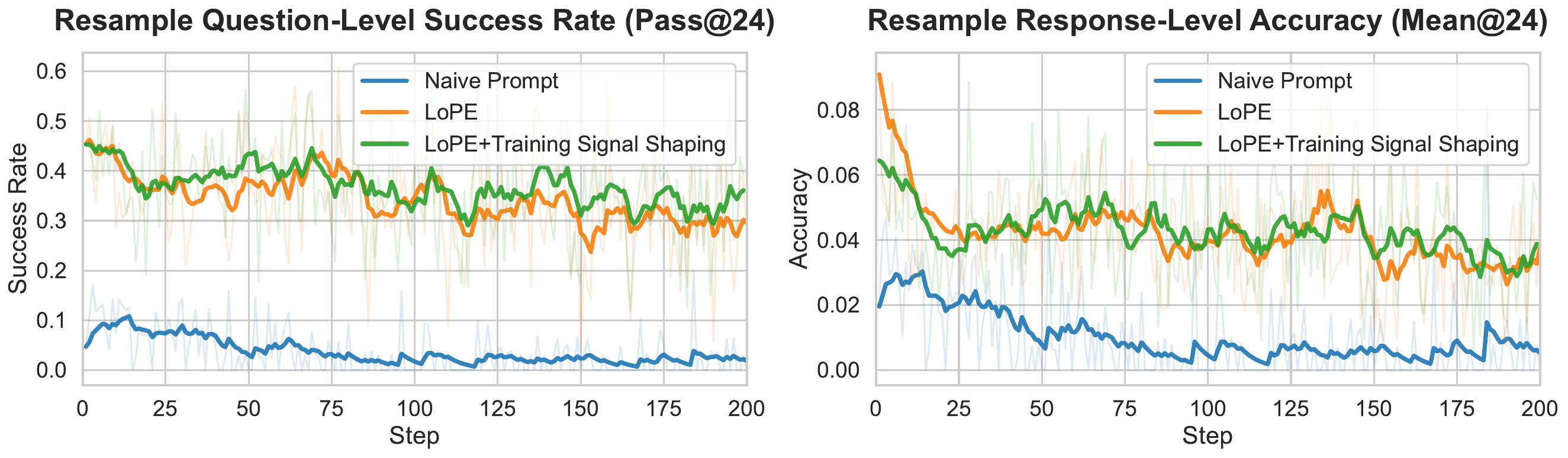}
    \caption{Resample success rate and accuracy during Qwen3-4B-Base training.}
    \vspace{-0.5em}
    \label{fig:training_qwen3_4b}
\end{figure}

\begin{figure}[h]
    \centering
    \includegraphics[width=\linewidth]{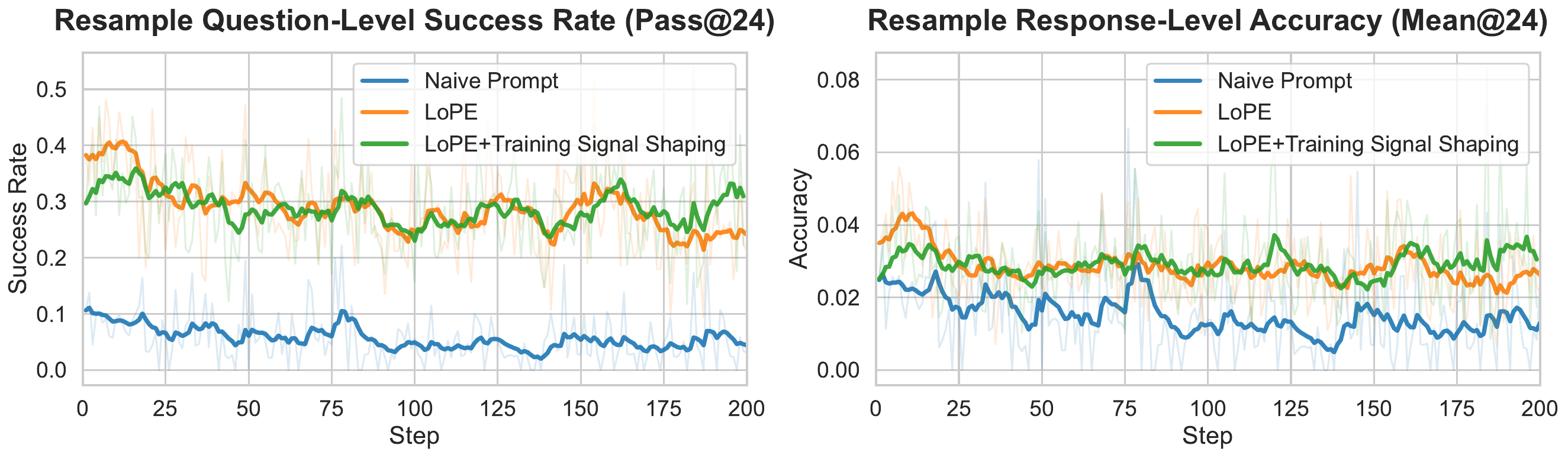}
    \caption{Resample success rate and accuracy during Qwen2.5-MATH-7B training.}
    \vspace{-0.5em}
    \label{fig:training_qwen25_7b}
\end{figure}

We present the training-time resample success rate and accuracy of Qwen3-4B-Base and Qwen2.5-Math-7B in Figure~\ref{fig:training_qwen3_4b} and Figure~\ref{fig:training_qwen25_7b}, where \method
consistently achieves a significantly higher question-level success rate.

\section{The Effectiveness of Training Signal Shaping on Training}
\label{appendix:tss_math}
\subsection{Policy Shaping}
\label{appendix:policyshape_math}
\paragraph{Gradient Analysis of the Resampling Objective.}
To better understand the role of the reward shaping function $f(\cdot)$ in our setting, we follow the derivation in \cite{yan2025learningreasonoffpolicyguidance} to analyze the gradient of the resampling part of the training objective in Eq.~\eqref{eq:lope}. Notably, unlike their formulation, we relax the assumption that $\pi_{\theta_{\text{old}}} = 1$. The objective is given by:
\begin{equation}
J_{\text{resample}}(\theta) = \mathbb{E}_{q,\,\{o_i\}\sim\pi_{\theta_{\text{old}}}} \sum_{i=1}^{N_s}\frac{1}{|o_i|}\sum_{t=1}^{|o_i|} \Big[f(\rho_{i,t})\,\hat{A}_i\Big],
\end{equation}

where $\rho_{i,t} = \pi_\theta / \pi_{\theta_{\text{old}}}$ is the importance sampling ratio. For brevity, we omit the conditioning variables in the following derivation and denote $\pi_\theta := \pi_\theta(o_{i,t}\mid q, o_{i,<t})$ and $\pi_{\theta_{\text{old}}} := \pi_{\theta_{\text{old}}}(o_{i,t}\mid \delta\oplus p, q, o_{i,<t})$.

% \paragraph{Gradient with respect to $\theta$.}
% Applying the logarithmic derivative $\nabla_\theta \pi_\theta = \pi_\theta \nabla_\theta \log \pi_\theta$, the derivative of $f(\rho_{i,t})$ w.r.t. $\theta$ is $f'(\rho_{i,t})\rho_{i,t} \nabla_\theta \log \pi_\theta$. Thus, the gradient of the resampling objective is:
% \begin{equation}
% \nabla_\theta J_{\text{resample}}(\theta) = \mathbb{E}_{q,\,\{o_i\}\sim\pi_{\theta_{\text{old}}}} \left[ \sum_{i,t} \frac{1}{|o_i|} \underbrace{f'(\rho_{i,t})}_{\text{shaped weight}} \rho_{i,t} \nabla_\theta \log \pi_\theta \cdot \hat{A}_i \right].
% \label{eq:resample-grad}
% \end{equation}
% The term $f'(\rho_{i,t})$ acts as a weighting function of the gradient. When $f(x)=x$, we have $f'(x)=1$, degenerating to the vanilla importance sampling gradient.

% \paragraph{Per-token Logit Gradient and Upper Bound.}
% We decompose the log probability and derive the gradient on each output logit $M_\theta(\tau')$, where $\tau'$ is any possible token in the action space:
% \begin{equation}
% \begin{aligned}
% \frac{\partial J_{\text{resample}}(\theta)}{\partial M_\theta(\tau')} &= \mathbb{E}_{q,\,\{o_i\}\sim\pi_{\theta_{\text{old}}}} \left[ f'(\rho_{i,t}) \rho_{i,t} \left( \mathbb{1}[\tau'=o_{i,t}] - \pi_\theta(\tau') \right) \cdot \hat{A}_i \right] \\
% \Rightarrow \left| \frac{\partial J_{\text{resample}}(\theta)}{\partial M_\theta(o_{i,t})} \right| &\le \mathbb{E}_{q,\,\{o_i\}\sim\pi_{\theta_{\text{old}}}} \left[ |f'(\rho_{i,t})| \rho_{i,t} (1 - \pi_\theta) \cdot |\hat{A}_i| \right],
% \end{aligned}
% \label{eq:logit-grad-bound}
% \end{equation}
\paragraph{Gradient with respect to $\theta$.}
Applying the logarithmic derivative $\nabla_\theta \pi_\theta = \pi_\theta \nabla_\theta \log \pi_\theta$, the derivative of $f(\rho_{i,t})$ w.r.t.\ $\theta$ is $f'(\rho_{i,t})\,\rho_{i,t}\,\nabla_\theta \log \pi_\theta$. Thus, the gradient of the resampling objective is:
\begin{equation}
\nabla_\theta J_{\text{resample}}(\theta) = \mathbb{E}_{q,\,\{o_i\}\sim\pi_{\theta_{\text{old}}}} \left[ \sum_{i,t} \frac{1}{|o_i|} \underbrace{f'(\rho_{i,t})}_{\text{shaped weight}} \rho_{i,t}\, \nabla_\theta \log \pi_\theta \cdot \hat{A}_i \right].
\label{eq:resample-grad}
\end{equation}
The term $f'(\rho_{i,t})$ acts as a weighting function of the gradient. When $f(x)=x$, we have $f'(x)=1$, degenerating to the vanilla importance sampling gradient.

\paragraph{Per-token Logit Gradient and Upper Bound.}
To analyze how the resampling objective updates the model's predictive distribution at the token level, we project the parameter gradient $\nabla_\theta \log \pi_\theta(o_{i,t})$ onto an individual output logit $M_\theta(\tau)$, where $\tau$ ranges over the action space at position $t$. Since the policy is parameterized as a softmax over logits,
\begin{equation}
\pi_\theta(o_{i,t}) = \frac{\exp\bigl(M_\theta(o_{i,t})\bigr)}{\sum_{\tau'} \exp\bigl(M_\theta(\tau')\bigr)},
\end{equation}
taking logarithms gives $\log \pi_\theta(o_{i,t}) = M_\theta(o_{i,t}) - \log \sum_{\tau'} \exp\bigl(M_\theta(\tau')\bigr)$, and differentiating with respect to $M_\theta(\tau)$ yields the standard softmax Jacobian identity:
\begin{equation}
\frac{\partial \log \pi_\theta(o_{i,t})}{\partial M_\theta(\tau)} = \mathbb{1}[\tau = o_{i,t}] - \pi_\theta(\tau).
\label{eq:softmax-jacobian}
\end{equation}
By taking the derivative of Eq.~\eqref{eq:resample-grad} with respect to $M_\theta(\tau)$ and applying Eq.~\eqref{eq:softmax-jacobian}, we obtain the per-token logit gradient and its upper bound:
\begin{equation}
\begin{aligned}
\frac{\partial J_{\text{resample}}(\theta)}{\partial M_\theta(\tau)} &= \mathbb{E}_{q,\,\{o_i\}\sim\pi_{\theta_{\text{old}}}} \left[ f'(\rho_{i,t})\, \rho_{i,t} \left( \mathbb{1}[\tau = o_{i,t}] - \pi_\theta(\tau) \right) \cdot \hat{A}_i \right] \\
\Rightarrow \left| \frac{\partial J_{\text{resample}}(\theta)}{\partial M_\theta(o_{i,t})} \right| &\le \mathbb{E}_{q,\,\{o_i\}\sim\pi_{\theta_{\text{old}}}} \left[ \bigl|f'(\rho_{i,t})\bigr|\, \rho_{i,t}\, (1 - \pi_\theta) \cdot \bigl|\hat{A}_i\bigr| \right],
\end{aligned}
\label{eq:logit-grad-bound}
\end{equation}

where the upper bound corresponds to the identity case $\tau = o_{i,t}$, yielding $|\mathbb{1}[\tau = o_{i,t}] - \pi_\theta(\tau)| = 1 - \pi_\theta$. This case captures the dominant gradient signal that elevates the logit of the resampled token itself. From Eq.~\eqref{eq:logit-grad-bound}, under the vanilla setting $f(x)=x$, the gradient scale is bounded by $\rho_{i,t}(1-\pi_\theta) = \pi_\theta(1-\pi_\theta)/\pi_{\theta_{\text{old}}}$.

\begin{figure}
    \centering
    \includegraphics[width=\linewidth]{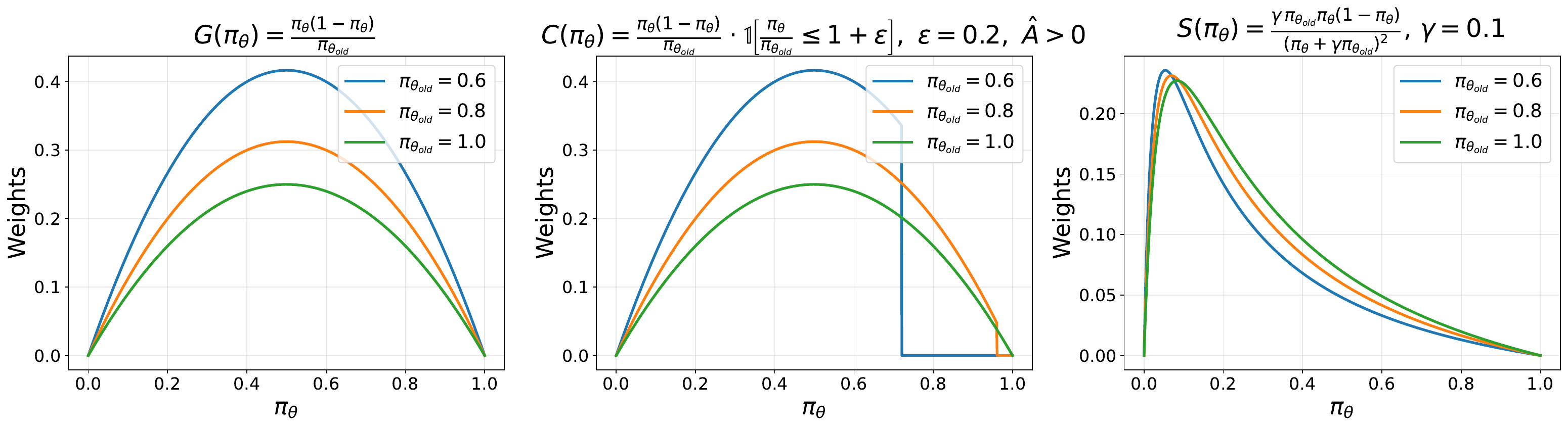}
    \caption{Per-token gradient weight under three formulations, plotted as a function of $\pi_\theta$.
\textbf{Left:} Vanilla gradient bound $G(\pi_\theta)$.
\textbf{Middle:} GRPO-clipped gradient $C(\pi_\theta)$ under positive advantage ($\hat{A}>0$). Gradients are truncated to zero when $\rho_{i,t}>1+\epsilon$.
\textbf{Right:} Policy-shaped gradient $S(\pi_\theta)$. The peak is relocated to the low-probability regime, with bounded peak value $1/4$.}
    \label{fig:gradient_compare}
\end{figure}

To visualize this bound and motivate our shaping choice, Figure~\ref{fig:gradient_compare} plots the per-token gradient weight as a function of $\pi_\theta$ across several $\pi_{\theta_{\text{old}}}$ values, under the situation of a positive advantage ($\hat{A}>0$). The left panel illustrates the unclipped vanilla gradient bound $G(\pi_{\theta})=\pi_\theta(1-\pi_\theta)/\pi_{\theta_{\text{old}}}$, while the middle panel displays its GRPO-clipped counterpart $C(\pi_{\theta})$. In $C(\pi_{\theta})$, the gradient is additionally truncated to zero when $\rho_{i,t}>1+\epsilon$, which corresponds to the active branch in the GRPO objective for $\hat{A}>0$.

As illustrated in Figure~\ref{fig:gradient_compare}, both the vanilla and GRPO-clipped gradient bounds suffer from two fundamental limitations:

\textbf{(i) Vanishing gradient at low $\pi_\theta$.} The bound decays to zero as $\pi_\theta \to 0$, yielding weak learning signals for unfamiliar tokens where the current policy is highly uncertain. Crucially, the distribution discrepancy between the resampled trajectories $\{o_i\} \sim \pi_{\theta_{\text{old}}}$ and the training policy $\pi_\theta$ systematically drives $\pi_\theta$ to small values, suppressing gradients exactly where off-policy guidance is most needed. GRPO clipping fails to address this issue, as it only operates on the $\rho_{i,t}>1+\epsilon$ region. 

\textbf{(ii) Inappropriate handling of low $\pi_{\theta_{\text{old}}}$ tokens.} The peak value of the vanilla bound is $1/(4\pi_{\theta_{\text{old}}})$, which diverges as $\pi_{\theta_{\text{old}}}\to 0$. This induces excessive gradient magnitudes on rare tokens of the resampling policy, causing training instability. While GRPO clipping avoids this by hard-truncating the gradients to zero, it entirely sacrifices the learning signals associated with these tokens.

To overcome these limitations, we adopt the policy shaping function $f(x)=x/(x+\gamma)$ proposed by \cite{yan2025learningreasonoffpolicyguidance} to reshape the gradients.

\paragraph{Specialization to $f(x)=x/(x+\gamma)$.}
The derivative of the shaping function is $f'(x)=\gamma/(x+\gamma)^2$. Substituting this into the bound in Eq.~\eqref{eq:logit-grad-bound} yields:
\begin{equation}
\left| \frac{\partial J_{\text{resample}}(\theta)}{\partial M_\theta(o_{i,t})} \right| \le \mathbb{E} \left[ \frac{\gamma\, \pi_{\theta_{\text{old}}}\, \pi_\theta (1-\pi_\theta)}{(\pi_\theta + \gamma\, \pi_{\theta_{\text{old}}})^2} \cdot |\hat{A}_i| \right].
\label{eq:shaped-bound}
\end{equation}
Treating the integrand of Eq.~\eqref{eq:shaped-bound} as a function of $\pi_\theta$ for a fixed $\pi_{\theta_{\text{old}}}$, it attains its maximum at:
\begin{equation}
\pi_\theta^{\star} \;=\; \frac{\gamma\, \pi_{\theta_{\text{old}}}}{1 + 2\gamma\, \pi_{\theta_{\text{old}}}}, \qquad
\text{with a peak value of} \quad
\frac{1}{4\bigl(1 + \gamma\, \pi_{\theta_{\text{old}}}\bigr)}.
\label{eq:peak}
\end{equation}
Equation~\eqref{eq:peak} formalizes two complementary properties of the shaping function, which are visually corroborated in Figure~\ref{fig:gradient_compare} (right). 

\textbf{(i) Low-probability emphasis.} The peak location shifts from $\pi_\theta = 1/2$ (the vanilla case) into the low-probability regime at $\pi_\theta^{\star} \!\approx\! \gamma\, \pi_{\theta_{\text{old}}}$. Consequently, the shaping function amplifies the learning signal for the unfamiliar yet highly rewarded tokens that off-policy resampling aims to introduce. 

\textbf{(ii) Bounded and stable peak value.} The peak value $1/[4(1+\gamma\,\pi_{\theta_{\text{old}}})]$ is strictly bounded by $1/4$ for all $\pi_{\theta_{\text{old}}}\!\in\![0,1]$ and remains stable across different $\pi_{\theta}$ and $\pi_{\theta_{old}}$ values. In contrast, the vanilla bound scales as $1/(4\pi_{\theta_{\text{old}}})$ and can grow unboundedly as $\pi_{\theta_{\text{old}}}\!\to\!0$. 

By reshaping the learning signal toward under-confident tokens and stabilizing the gradient magnitude across the full spectrum of probabilities, policy shaping effectively reweights the parameter updates. It assigns greater importance to low-probability yet effective actions while gracefully attenuating updates for tokens the model has already mastered.

\subsection{Advantage Shaping}
\label{appendix:advshape_math}

\begin{figure}[h]
    \centering
    \includegraphics[width=\linewidth]{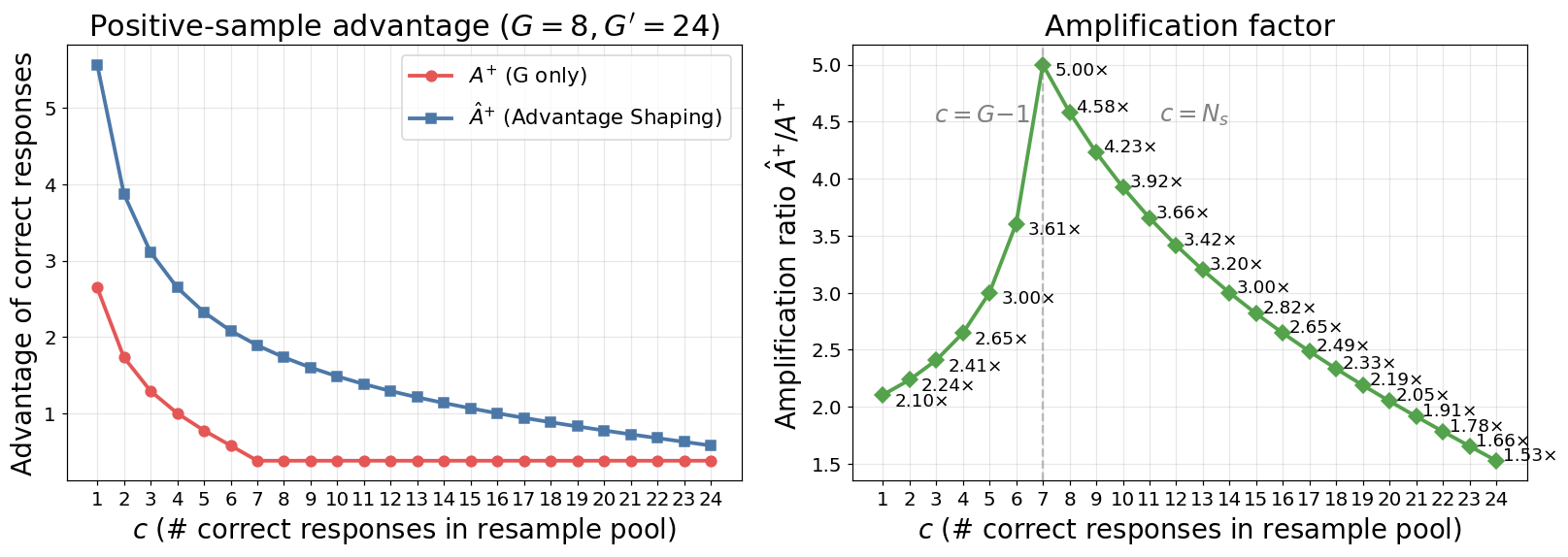}
    \caption{Comparison of advantages for positive responses before and after advantage shaping. \textbf{Left:} The absolute advantage values of the vanilla advantage ($A^+$) and shaped advantage ($\hat{A}^+$), as a function of the number of correct responses $c$. \textbf{Right:} The amplification factor ($\hat{A}^+ / A^+$) achieved by applying advantage shaping.}
    \label{fig:advantage_shaping}
\end{figure}

\paragraph{Quantitative Effect of Advantage Shaping.}
Since the advantage term acts as a constant multiplier in the training objective (Eq.~\ref{eq:lope}) and the gradient (Eq.~\ref{eq:resample-grad}), its effect is independent of other variables. Therefore, we quantitatively analyze the advantage term in isolation.

We follow the notations in Sections~\ref{sec:lope} and \ref{sec:shape}: the rollout size for the first-time sampling and the resampling is $G$ and $G'$, where the number of correct responses is 0 and $c$, respectively. The gradient update is performed on a regrouped set of $G$ responses, consisting of $N_s = \min(c, G-1)$ correct resampled responses and $G-N_s$ incorrect first-time sampling responses. The vanilla \method calculates the advantage, denoted as $A$, solely within the $G$ selected rollouts. In contrast, advantage shaping computes the advantage, denoted as $\hat{A}$, over the entire pool of $G+G'$ rollouts.

In these two scenarios, the means and standard deviations of the rewards are $\mu=N_s/G,\ \sigma=\sqrt{(G-N_s)N_s}/G$ and $\hat{\mu}=c/(G+G'),\ \hat{\sigma}=\sqrt{(G+G'-c)c}/(G+G')$, respectively. Substituting these into Eq.~\ref{eq:standard_adv} and Eq.~\ref{eq:lope_adv}, we obtain the advantages for the positive samples:
\begin{equation}
A^{+} = \sqrt{\frac{G - N_s}{N_s}}, \qquad
\hat{A}^{+} = \sqrt{\frac{(G+G') - c}{c}}.
\label{eq:advantage-closed-form}
\end{equation}

Figure~\ref{fig:advantage_shaping} visualizes these quantities and their ratio using our actual sampling budget ($G=8, G'=24$) across the regime of $c \in [1, G']$. Since resampling is triggered exclusively on hard questions where the initial $G$ samples all fail, \textbf{$c$ is typically smaller than $G$ in practice ($c<G$)}. With in this range, the amplification factor $\hat{A}^{+}/A^{+}$ ranges from $2.10\times$ at $c=1$ to a peak of $5.00\times$ at $c=G-1=7$. 

According to Eq.~\eqref{eq:resample-grad}, the amplification of advantage directly applies to the gradient weight. Therefore, advantage shaping effectively assigns a larger training weight to the rare correct trajectories that drive learning on hard questions.

\end{document}